\documentclass[accepted]{uai2025} % after acceptance, for a revised version; 
\usepackage[american]{babel}
\usepackage{natbib}
\usepackage{mathtools} % amsmath with fixes and additions
\usepackage{booktabs} % commands to create good-looking tables
\usepackage{tikz} % nice language for creating drawings and diagrams

\usepackage{authblk}
\usepackage{algorithm}
\usepackage{algorithmic}
\usepackage{multirow}
\usepackage{subfigure}
\usepackage{amsmath}
\usepackage{amssymb}
\usepackage{amsthm}
\newtheorem{theorem}{Theorem}
\usepackage{bbm}
\usepackage{bm}
\usepackage{setspace}
\usepackage{url}
\usepackage{graphicx}
\usepackage[toc,page]{appendix}

\newtheorem{innercustomthm}{Theorem}
\newenvironment{customthm}[1]
{\renewcommand\theinnercustomthm{#1}\innercustomthm}{\endinnercustomthm}

\title{Label Distribution Learning \\using the Squared Neural Family on the Probability Simplex}

\author[1,2]{Daokun Zhang}
\author[2]{Russell Tsuchida}
\author[3]{Dino Sejdinovic}
\affil[1]{%
    School of Computer Science\\
    University of Nottingham Ningbo China
}
\affil[2]{%
    Department of Data Science \& AI\\
    Monash University
}
\affil[3]{%
    School of Computer and Mathematical Sciences\\
    The University of Adelaide
}

\begin{document}
\onecolumn
\maketitle
\begin{abstract}
Label distribution learning (LDL) provides a framework wherein a distribution over categories rather than a single category is predicted, with the aim of addressing ambiguity in labeled data.
Existing research on LDL mainly focuses on the task of point estimation, i.e., finding an optimal distribution in the probability simplex conditioned on the given sample. 
In this paper, we propose a novel label distribution learning model SNEFY-LDL, which estimates a probability distribution of all possible label distributions over the simplex, by unleashing the expressive power of the recently introduced Squared Neural Family (SNEFY), a new class of tractable probability models. %in modeling complicated probability distributions. 
%With the modeled distribution, label distribution prediction can be achieved by performing the expectation operation to estimate the mean of the distribution of label distributions. 
As a way to summarize the fitted model, we derive the closed-form label distribution mean, variance and covariance conditioned on the given sample, which can be used to predict the ground-truth label distributions, construct label distribution confidence intervals, and measure the correlations between different labels.
Moreover, more information about the label distribution prediction uncertainties can be acquired from the modeled probability density function. 
Extensive experiments on conformal prediction, active learning and ensemble learning are conducted, verifying SNEFY-LDL's great effectiveness in LDL uncertainty quantification. 
%The promising experimental results demonstrate the great utility of SNEFY-LDL in a broad range of uncertainty-aware label distribution learning applications. 
The source code of this paper is available at \url{https://github.com/daokunzhang/SNEFY-LDL}.

%the label distribution prediction task, showing that our distribution modeling based method can achieve very competitive label distribution prediction performance compared with the state-of-the-art baselines. 
%Additional experiments on active learning and ensemble learning demonstrate that our probabilistic approach can effectively boost the performance in these settings, by accurately estimating the prediction reliability and uncertainties. 
\end{abstract}

\section{Introduction}\label{sec:intro}

Label distribution learning (LDL) is a technique which handles ambiguity in multi-class classification, by utilizing simplex-valued rather than categorical-valued labels in training data.
Unlike traditional multi-class and multi-label learning paradigms, which assign a deterministic label prediction to instances, LDL corresponds to the question "\textit{How well does each of the labels describe an instance?}", by using a discrete probability distribution to characterize each label's composition ratio in jointly describing the given instance. 
For example, when we predict the functionality of a district in a city, we might predict a result such as: the district has 20\% functionality for business, 40\% functionality for entertainment, and 40\% functionality for education. 

Many LDL algorithms have been proposed to directly predict label distribution vectors from instance features, by adapting machine learning algorithms designed for "hard" label prediction to the "soft" label prediction setting. 
Though a discrete distribution among candidate labels is predicted, existing LDL algorithms still operate at the level of point estimation, i.e., they search for a single point on a probability simplex (the set of all possible label distributions) for each given instance. 
The point estimation paradigm is particularly susceptible to data uncertainty and inexact mappings between instances and labels, due to the inherent complexity of the data collection and generation processes. 
Therefore, modeling the probability distribution of label distribution vectors, i.e., the probability distribution supported on the probability simplex, is an important step towards trustworthy LDL.  
An additional bonus of the distribution modeling is the ability to quantify the prediction reliability and uncertainties, which not only facilitates reliable model deployment in real-world safety critical applications, but is also essential to various reliability/uncertainty-aware tasks, like pseudo labeling, active learning, and ensemble learning. 

\textbf{Contributions.} In this paper, we propose a novel LDL framework, SNEFY-LDL, by unleashing the probability modeling power of the recently introduced Squared Neural Family (SNEFY)~\citep{tsuchida2024squared}, a new class of tractable probability models.
%which models a probability distribution on label distribution vectors conditioned on the input sample. 
By restricting the support set of SNEFY to a probability simplex, 
%and employing the exponential function as the activation function, and lebsegue measure as the base measure, 
SNEFY-LDL constructs an expressive multimodal distribution modeling of the label distribution vector conditioned the given sample. The conditional distribution model has a closed-form normalizing constant, guaranteeing computational tractability. In this way, model parameters can be learned efficiently by maximizing the conditional likelihoods of training samples with stochastic gradient descent. 
As a way to summarize the fitted model, we derive the closed-form label distribution mean, variance and covariance conditioned on the given sample, which can be used to predict the ground-truth label distribution, construct label distribution confidence intervals, and measure the correlations between different labels.
%Through computing the expectation of the conditional distribution, we can obtain the closed-form distribution mean, as a way to summarize the fitted model and predict label distributions for unlabeled samples. 
However, 
%this is only one way to use the fitted model 
the fitted model is not limited to these usecases, 
and the probability density values can be used to directly evaluate the reliability of label distribution predictions.

%We conduct extensive experiments on fourteen real-world datasets to evaluate the label distribution prediction performance of the proposed SNEFY-LDL model. 
%Experimental results show that SNEFY-LDL can achieve very competitive label distribution prediction performance, compared with the state-of-the-art LDL baselines. 
We conduct extensive experiments on label distribution conformal prediction, active learning and ensemble learning to verify the efficacy of SNEFY-LDL in quantifying prediction uncertainties. 
For the task of conformal prediction, we use the SNEFY-LDL's closed-form conditional mean and variance of label distribution predictions to construct a confidence interval for each label's composition ratio in describing the given instance and calibrate the confidence interval through conformal prediction~\citep{angelopoulos2021gentle}. Experimental results show that the confidence intervals constructed by the SNEFY-LDL model have greater adaptivity than the confidence intervals constructed by the naive Dirichlet distribution. The max-entropy principle is used to achieve active learning with the estimated SNEFY-LDL entropy, i.e., select the most informative unlabeled samples with the largest entropy values, query their labels and augment training samples, to attain the largest performance gain of the re-trained LDL model.
Experimental results show that the max-entropy principle achieves significantly better active learning performance than the representativeness based active learning baselines. 
The experiments on ensemble learning demonstrate that SNEFY-LDL gives a further usecase for the fitted probabilistic model, as it provides an intelligent mechanism for weighting base learners, significantly outperforming the uniform weighting strategy. 

\section{Related Work}

LDL is first proposed by~\citet{geng2013facial} to solve the facial age estimation problem.
%with insufficient training samples. 
Since then, a series of LDL algorithms have been developed, which are mainly in three categories: Problem Transformation (PT), Algorithm Adaptation (AA) and Specialized Algorithms (SA)~\citep{geng2016label}. 

\textbf{Problem Transformation.} PT~\citep{geng2016label} transforms the LDL problem into the single-label classification problem, by decomposing each training sample assigned with a label distribution into a set of duplicate training samples. 
Each of them is assigned with a different label and accounts for a partial sample in proportion to the label probability value, and is then used to train the single-label classifiers. 
The label likelihoods predicted by the single-label classifiers are then aggregated to form the final prediction of label distributions. 
PT-Bayes~\citep{geng2016label} and PT-SVM~\citep{geng2015pre}  transform the LDL problem into the single-label multi-class classification problem and respectively employ Bayes and SVM as the single-label classifiers. DF-LDL~\citep{gonzalez2021decomposition} decomposes the label distribution prediction task into a number of one-versus-one binary classification tasks, and fuses the binary classification likelihoods to form the final label distribution predictions. 

\textbf{Algorithm Adaptation.} AA~\citep{geng2016label} adapts traditional single-label classification models into the LDL setting, by leveraging the models' compatibility in outputting a soft label distribution vector. 
Derived from the K Nearest Neighbor (KNN) algorithm~\citep{wu2008top}, AA-KNN~\citep{geng2016label} predicts samples' label distributions by averaging the label distributions of their $k$ nearest neighbors in feature space. 
AA-BP~\citep{geng2016label} constructs a three-layer neural network and adopts the softmax function as the activation of the output layer, making the neural network naturally produce a label distribution for each example. 
The neural network is trained by minimizing the sum of squared errors between the  model-output label distributions and the ground-truth label distributions.

\textbf{Specialized Algorithms.} SA~\citep{geng2016label} designs algorithms from scratch to directly solve the LDL problem. 
SA-IIS~\citep{geng2013facial} and SA-BFGS~\citep{geng2016label} use the maximum entropy model to parameterize label distributions. They are trained by minimizing the Kullback-Leibler (KL) divergence between the model-output and ground-truth label distributions, where Improved Iterative Scaling (IIS)~\citep{malouf2002comparison} and BFGS~\citep{nocedal1999numerical} are respectively leveraged by SA-IIS and SA-BFGS as optimizers. 
CPNN~\citep{geng2013facial} uses a neural network to parameterize the joint probability distribution between sample features and labels following Modha’s probability distribution
%density function (PDF) 
formulation~\citep{modha1994learning}.
BCPNN~\citep{yang2017learning} and ACPNN~\citep{yang2017learning} then improve on CPNN through leveraging binary label encoding and augmenting training samples respectively. 
LDLF~\citep{shen2017label} employs differentiable decision trees~\citep{kontschieder2015deep} to model label distributions and KL divergence is used to design the learning objective. 
LDL-SCL~\citep{jia2019label} forces the label distributions of samples located closely in feature space to be similar to each other. LDL-LRR~\citep{jia2021label} and LDL-DPA~\citep{jia2023adaptive} maintain the relative importance ranking between different labels in label distribution modeling, by penalizing a label importance ranking loss in their learning objectives.

\textbf{Extensions.} In addition, LDL has been extended to other tasks, like label enhancement 
%(recovering label distributions from the one-hot single-label assignments)
~\citep{xu2019label,xu2020variational,zheng2023label}, multi-class classification~\citep{wang2019classification,wang2021learn,wang2021label}, learning with incomplete supervision~\citep{xu2017incomplete}, oversampling~\citep{gonzalez2021synthetic}, ordinal LDL~\citep{wen2023ordinal}, semi-supervised learning~\citep{xie2023rankmatch}, and label calibration~\citep{he2024generative}. Furthermore, LDL has been applied to solve numerous real-world problems, including facial age estimation~\citep{geng2013facial}, facial emotion recognition~\citep{chen2020label}, head pose estimation~\citep{xu2019head}, crowd opinion prediction~\citep{geng2015pre}, emphasis selection~\citep{shirani2019learning}, lesion counting~\citep{wu2019joint}, and urban functionality prediction~\citep{huang2023learning}. It is worth noting that there is a related topic in the classical statistics literature, termed \emph{compositional data analysis} \citep{greenacre2021compositional}, with a broad range of applications including geochemistry (e.g. labels correspond to mineral compositions) and ecology (e.g. labels correspond to relative abundance of species).

However, existing LDL algorithms mainly fall into the regime of point estimation. 
They discover an optimal discrete label distribution in the probability simplex with regard to a predefined learning objective, and do not provide the information about how prominent the optimal label  distribution is, compared with the remaining distribution candidates. In this paper, we aim to model the distribution of label distributions, with the expectation that we can provide a promising mechanism for LDL uncertainty quantification.

\section{Problem Definition}
Assume we are given a set of $N$ i.i.d training samples $\mathcal{X}=\{\bm{x}_1, \bm{x}_2,\cdots,\bm{x}_{N}\}$ with each sample $\bm{x}\in\mathcal{X}$ located in the $d$-dimensional Euclidean space $\mathbb{R}^d$. In addition, each sample $\bm{x}\in\mathcal{X}$ is described by  a $L$-dimensional label distribution vector $\bm{\ell}_{\bm{x}}\in\mathbb{R}^{L}$ that takes values in the $(L-1)$-simplex $\Delta^{L-1}$, corresponding to a set of $L$ given labels $\mathcal{Y}=\{y_1,y_2,\cdots,y_{L}\}$. The $l$-th entry $\bm{\ell}_{\bm{x}}^{y_l}$ of the label distribution vector $\bm{\ell}_{\bm{x}}$ corresponds to the composition ratio of the $l$-th label $y_l$ in describing $\bm{x}$, 
%the probability of sample $\bm{x}$ belonging to the $l$-th class $y_l$, i.e., $\bm{\ell}_{\bm{x}}^{y_l}=\mathrm{P}(y_l|\bm{x}) \geq 0$, 
satisfying the constraint that $\sum_{l=1}^{L}\bm{\ell}_{\bm{x}}^{y_l}=1$.

With the label distribution observations of training samples, our objective is to model the probability distribution of the label distribution vector $\bm{\ell}\in\Delta^{L-1}$ conditioned on any input sample $\bm{x}\in\mathbb{R}^{d}$,  $\mathrm{P}(d\bm{\ell}|\bm{x})$. 

%For any sample $\bm{x}\in\mathbb{R}^{d}$ with an unknown label distribution, one can associate to it a label distribution vector $\bm{\ell}^{*}_{\bm{x}}$ obtained as the mean of the fitted distribution $\mathrm{P}(d\bm{\ell}|\bm{x})$:
%\begin{equation}
%\bm{\ell}^{*}_{\bm{x}}=\int_{\Delta^{L-1}}\bm{\ell} \, \mathrm{P}(d\bm{\ell}|\bm{x}),
%\end{equation}
%but the fitted model $\mathrm{P}(d\bm{\ell}|\bm{x})$ can serve many other purposes.

\section{Preliminaries on SNEFY}

Given a measure space $(\Omega,\mathcal{F},\mu)$ with set $\Omega$, sigma algebra $\mathcal{F}$, and nonnegative measure $\mu$, SNEFY defines a probability distribution $\mathrm{P}$ on some support 
%\textcolor{red}{The symbol $\mathbb{Z}$ maybe clashes with the usual set of integers.} 
$\mathcal{Z}\in\mathcal{F}$ %with respect to the base measure $\mu$, 
to be proportional to the evaluation of the squared 2-norm of a neural network $\bm{f}$:
\begin{equation}
\begin{aligned}
&\mathrm{P}(d\bm{z};\bm{V},\bm{\Theta})\triangleq\frac{{\|\bm{f}(\bm{t}(\bm{z});\bm{V},\bm{\Theta})\|}^2_2\mu(d\bm{z})}{\int_{\mathcal{Z}}{\|\bm{f}(\bm{t}(\bm{z});\bm{V},\bm{\Theta})\|}^2_2\mu(d\bm{z})},\\
&\bm{f}(\bm{t}(\bm{z});\bm{V},\bm{\Theta})=\bm{V}\sigma(\bm{W}\bm{t}(\bm{z})+\bm{b}),\;\bm{\Theta}=(\bm{W},\bm{b}),
\end{aligned}
\label{eq:dist_formula_1}
\end{equation}
where $\bm{t}(\cdot): \mathcal{Z}\rightarrow\mathbb{R}^D$ is the sufficient statistic, $\sigma$ is the activation function,  $\bm{W}\in\mathbb{R}^{n\times D}$ and $\bm{b}\in\mathbb{R}^{n}$ are respectively the weight matrix and bias vector at the hidden layer of neural network $\bm{f}$, and $\bm{V}\in\mathbb{R}^{m\times n}$ are $\bm{f}$'s readout parameters. 
%\textcolor{red}{Something wrong with shapes here. $\bm{V}$ maybe should be $m \times n$?}
$\bm{\Theta}=(\bm{W},\bm{b})\in\mathbb{R}^{n\times(D+1)}$ is the concatenation of $\bm{W}$ and $\bm{b}$ and its $i$th row is denoted as $\bm{\theta}_i=(\bm{w}_i,b_i)\in\mathbb{R}^{D+1}$, where $\bm{w}_i\in\mathbb{R}^D$ is the $i$th row of $\bm{W}$ and $b_i$ is the $i$th element of $\bm{b}$.

The distribution $\mathrm{P}(d\bm{z};\bm{V},\bm{\Theta})$ in Eq. (\ref{eq:dist_formula_1}) admits a more concise formulation
\begin{equation}
\begin{aligned}
\mathrm{P}(d\bm{z};\bm{V},\bm{\Theta})&=\frac{\mathrm{Tr}[\bm{V}^{\top}\bm{V}\widetilde{\bm{K}}_{\bm{\Theta}}(\bm{z})]}{\mathrm{Tr}[\bm{V}^{\top}\bm{V}\bm{K}_{\bm{\Theta}}]}\mu(d\bm{z}),\\
&=\frac{\mathrm{vec}(\bm{V}^{\top}\bm{V})^{\top}\mathrm{vec}(\widetilde{\bm{K}}_{\bm{\Theta}}(\bm{z}))}{\mathrm{vec}(\bm{V}^{\top}\bm{V})^{\top}\mathrm{vec}(\bm{K}_{\bm{\Theta}})}\mu(d\bm{z}),
\end{aligned}
\label{eq:dist_formula_2}
\end{equation}
where $\widetilde{\bm{K}}_{\bm{\Theta}}(\bm{z})\in\mathbb{R}^{n\times n}$ is a positive semidefinite (PSD) matrix, whose $ij$th element is a kernel function of $\bm{\theta}_i$ and $\bm{\theta}_j$:
\begin{equation}
\tilde{\bm{k}}_{\sigma,\bm{t}}(\bm{\theta}_i,\bm{\theta}_j;\bm{z})=\sigma(\bm{w}_i^{\top}\bm{t}(\bm{z})+b_i)\sigma(\bm{w}_j^{\top}\bm{t}(\bm{z})+b_j),
\end{equation}
while $\bm{K}_{\bm{\Theta}}$ is the elementwise integral of $\widetilde{\bm{K}}_{\bm{\Theta}}(\bm{z})$, preserving the PSD property, with its $ij$th entry formulated as another kernel function of $\bm{\theta}_i$ and $\bm{\theta}_j$:
\begin{equation}
\bm{k}_{\sigma,\bm{t},\mu}(\bm{\theta}_i,\bm{\theta}_j)=\int_{\mathcal{Z}}\tilde{\bm{k}}_{\sigma,\bm{t}}(\bm{\theta}_i,\bm{\theta}_j;\bm{z})\mu(d\bm{z}).
\end{equation}
Under varying choices of the activation function $\sigma$, sufficient statistic $\bm{t}$, and the base measure $\mu$, $\bm{k}_{\sigma,\bm{t},\mu}(\bm{\theta}_i,\bm{\theta}_j)$ is able to be computed in closed form (see Table~1 of~\citep{tsuchida2024squared}) in $\mathcal{O}(D)$.
%$\mathcal{O}(m^2n + d n^2)$. 
This makes SNEFY a tractable probability distribution model, with great expressivity and computational efficiency. 

SNEFY also enjoys a closed-form formulation for conditional distributions, under mild conditions. 
\begin{theorem}~\citep{tsuchida2024squared}
Let $\mathtt{z}=(\mathtt{z}_1,\mathtt{z}_2)$ jointly follow a SNEFY distribution, with support set $\mathcal{Z}=\mathcal{Z}_1\times\mathcal{Z}_2$, sufficient statistic $\bm{t}$, activation function $\sigma$, base measure $\mu$, as well as parameters $\bm{V}$ and $\bm{\Theta}=([\bm{W}_1,\bm{W}_2],\bm{b})$. Assume that $\mu(d\bm{z})=\mu_1(d\bm{z}_1)\mu_2(d\bm{z}_2)$ and $\bm{t}(\bm{z})=(\bm{t}_1(\bm{z}_1),\bm{t}_2(\bm{z}_2))$. Then the conditional distribution of $\mathtt{z}_1$ given $\mathtt{z}_2=\bm{z}_2$ is also a SNEFY distribution with support set $\mathcal{Z}_1$, sufficient statistic $\bm{t}_1$, activation function $\sigma$, base measure $\mu_1$, as well as parameters $\bm{V}$ and $\bm{\Theta}_{1|2}=(\bm{W}_1,\bm{W}_2\bm{t}_2(\bm{z}_2)+\bm{b})$.
\end{theorem}

\section{SNEFY-LDL}
SNEFY provides an effective way to model the conditional distribution of the label distribution vector $\bm{\ell}\in\Delta^{L-1}$. 
We can assume that the concatenation of label distribution vector $\bm{\ell}\in\Delta^{L-1}$ and its conditioning sample $\bm{x}\in\mathbb{R}^{d}$, $\bm{z}=(\bm{\ell},\bm{x})$, follows a joint SNEFY distribution, with support set $\mathcal{Z}=\Delta^{L-1}\times\mathbb{R}^d$, sufficient statistic $\bm{t}(\bm{z})=(\bm{t}_1(\bm{\ell}),\bm{t}_2(\bm{x})) : \mathcal{Z}\rightarrow\mathbb{R}^{D_1+D_2}$ composed of  $\bm{t}_1(\cdot):\Delta^{L-1}\rightarrow\mathbb{R}^{D_1}$ and $\bm{t}_2(\cdot):\mathbb{R}^d\rightarrow\mathbb{R}^{D_2}$, activation function $\sigma$, base measure $\mu(\bm{z})=\mu_1(\bm{\ell})\mu_2(\bm{x})$, as well as parameters $\bm{V}\in\mathbb{R}^{m\times n}$ and $\bm{\Theta}=([\bm{W}_1,\bm{W}_2],\bm{b})\in\mathbb{R}^{n\times(D_1+D_2+1)}$. Following \textbf{Theorem 1}, given sample $\bm{x}$, the conditional distribution of its label distribution vector $\bm{\ell}$ is a SNEFY distribution with support set $\Delta^{L-1}$, sufficient statistic $\bm{t}_1$, activation function $\sigma$, base measure $\mu_1$, as well as parameters $\bm{V}$ and $\bm{\Theta}_{1|2}=(\bm{W}_1, \bm{W}_2\bm{t}_2(\bm{x})+\bm{b})$. The conditional distribution is
\begin{equation}
\begin{aligned}
&\mathrm{P}(d\bm{\ell}|\bm{x};\bm{V},\bm{\Theta})\triangleq\frac{{\|\bm{f}(\bm{t}_1(\bm{\ell}),\bm{t}_2(\bm{x});\bm{V},\bm{\Theta})\|}^2_2\mu_1(d\bm{\ell})}{\int_{\Delta^{L-1}}{\|\bm{f}(\bm{t}_1(\bm{\ell}),\bm{t}_2(\bm{x});\bm{V},\bm{\Theta})\|}^2_2\mu_1(d\bm{\ell})},\\
&\bm{f}\big(\bm{t}_1(\bm{\ell}),\bm{t}_2(\bm{x});\bm{V},\bm{\Theta}\big)=\bm{V}\sigma\big(\bm{W}_1\bm{t}_1(\bm{\ell})+\bm{W}_2\bm{t}_2(\bm{x})+\bm{b}\big).\\
\end{aligned}
\end{equation}
Following Eq.~\eqref{eq:dist_formula_2}, the distribution 
%$\mathrm{P}(d\bm{\ell}|\bm{x};\bm{V},\bm{\Theta})$ 
can be reformulated as
\begin{equation}
\begin{aligned}
\mathrm{P}(d\bm{\ell}|\bm{x};\bm{V},\bm{\Theta})&=\frac{\mathrm{Tr}[\bm{V}^{\top}\bm{V}\widetilde{\bm{K}}_{\bm{\Theta}}(\bm{\ell},\bm{x})]}{\mathrm{Tr}[\bm{V}^{\top}\bm{V}\bm{K}_{\bm{\Theta}}(\bm{x})]}\mu_1(d\bm{\ell}),\\
&=\frac{\mathrm{vec}(\bm{V}^{\top}\bm{V})^{\top}\mathrm{vec}(\widetilde{\bm{K}}_{\bm{\Theta}}(\bm{\ell},\bm{x}))}{\mathrm{vec}(\bm{V}^{\top}\bm{V})^{\top}\mathrm{vec}(\bm{K}_{\bm{\Theta}}(\bm{x}))}\mu_1(d\bm{\ell}), \label{eq:snefy_inner}
\end{aligned}
\end{equation}
where $\widetilde{\bm{K}}_{\bm{\Theta}}(\bm{\ell},\bm{x})\in\mathbb{R}^{n\times n}$ is a PSD matrix, with its $ij$th element being a kernel function of $\bm{\theta}_i=(\bm{w}_{1i},\bm{w}_{2i},b_i)\in\mathbb{R}^{D_1+D_2+1}$ and $\bm{\theta}_j=(\bm{w}_{1j},\bm{w}_{2j},b_j)\in\mathbb{R}^{D_1+D_2+1}$:
\begin{equation}
%\begin{aligned}
\tilde{\bm{k}}_{\sigma,\bm{t}_1,\bm{t}_2}(\bm{\theta}_i,\bm{\theta}_j;\bm{\ell},\bm{x})=
\sigma(\bm{w}_{1i}^{\top}\bm{t}_1(\bm{\ell})+\bm{w}_{2i}^{\top}\bm{t}_2(\bm{x})+b_i)\cdot
\sigma(\bm{w}_{1j}^{\top}\bm{t}_1(\bm{\ell})+\bm{w}_{2j}^{\top}\bm{t}_2(\bm{x})+b_j),
%\end{aligned}
\end{equation}
where $\bm{w}_{1i}\in\mathbb{R}^{D_1}$ and $\bm{w}_{2i}\in\mathbb{R}^{D_2}$ are respectively the $i$th row of matrices $\bm{W}_1$ and $\bm{W}_2$, and $b_i$ is the $i$th element of the bias vector $\bm{b}$. Then $\bm{K}_{\bm{\Theta}}(\bm{x})$ is the elementwise integral of $\widetilde{\bm{K}}_{\bm{\Theta}}(\bm{\ell},\bm{x})$, preserving the PSD property, with its $ij$th entry formulated as another kernel function of $\bm{\theta}_i$ and $\bm{\theta}_j$:
\begin{equation}
\bm{k}_{\sigma,\bm{t}_1,\bm{t}_2,\mu_1}(\bm{\theta}_i,\bm{\theta}_j;\bm{x})=\int_{\Delta^{L-1}}\tilde{\bm{k}}_{\sigma,\bm{t}_1,\bm{t}_2}(\bm{\theta}_i,\bm{\theta}_j;\bm{\ell},\bm{x})\mu_1(d\bm{\ell}).
\label{eq:condition_kernel}
\end{equation}
By choosing the activation function $\sigma$, sufficient statistic $\bm{t}_1$ and the base measure $\mu_1$ carefully, the kernel function $\bm{k}_{\sigma,\bm{t}_1,\bm{t}_2,\mu_1}(\bm{\theta}_i,\bm{\theta}_j;\bm{x})$ admits a closed form, which guarantees that the conditional distribution $\mathrm{P}(d\bm{\ell}|\bm{x};\bm{V},\bm{\Theta})$ is tractable.
In particular, we have the following theorem:
\begin{theorem}
Let $\bm{t}_1(\bm{\ell})=(\log\bm{\ell}^{y_1},\log\bm{\ell}^{y_2},\cdots,\log\bm{\ell}^{y_L}):\Delta^{L-1}\rightarrow\mathbb{R}^{L}$ by setting $D_1=L$, the activation function $\sigma$ be the exponential function $\exp$, the base measure $\mu_1(d\bm{\ell})=d\bm{\ell}$ be the Lebesgue measure. Under the condition that $\bm{W}_1>-1/2$ elementwise, the kernel function $\bm{k}_{\sigma,\bm{t}_1,\bm{t}_2,\mu_1}(\bm{\theta}_i,\bm{\theta}_j;\bm{x})$ admits a closed form:
\begin{equation}
%\begin{aligned}
\bm{k}_{\bm{t}_2}(\bm{\theta}_i,\bm{\theta}_j;\bm{x})=\exp(\bm{w}_{2i}^{\top}\bm{t}_2(\bm{x})+\bm{w}_{2j}^{\top}\bm{t}_2(\bm{x})+b_i+b_j)\cdot
\frac{\prod_{l=1}^{L}\Gamma(1+w_{1il}+w_{1jl})}{\Gamma\big(L+\sum_{l=1}^{L}(w_{1il}+w_{1jl})\big)},
\label{eq:kernel_dirichlet}
%\end{aligned}
\end{equation}
where $w_{1il}$ is the $il$-th element of matrix $\bm{W}_1$ and $\Gamma(\cdot)$ is the gamma function.
\end{theorem}
\noindent The proof is provided in the Appendix.
With the closed-form kernel function in Eq.~\eqref{eq:kernel_dirichlet}, we can construct the conditional SNEFY distribution $\mathrm{P}(d\bm{\ell}|\bm{x};\bm{V},\bm{\Theta})$ in the form of Eq.~\eqref{eq:snefy_inner}. 
This model provides us with the freedom to choose any sufficient statistic $\bm{t}_2(\cdot)$ that is used to transform the input sample $\bm{x}$ from the original $d$-dimensional space to the latent $D_2$-dimensional space.  
To capture the non-linearity between input samples and their label distributions, deep neural networks can be leveraged to construct $\bm{t}_2(\cdot)$. 
The input can also be extended beyond the vector-format samples to data with special structures, like images, texts and graphs, where we can respectively leverage Convolutional Neural Networks (CNNs)~\citep{venkatesan2017convolutional},  Transformers~\citep{vaswani2017attention} and Graph Neural Networks (GNNs)~\citep{kipf2016semi} to construct $\bm{t}_2(\cdot)$ for end-to-end learning.

The conditional distribution formulation $\mathrm{P}(d\bm{\ell}|\bm{x};\bm{V},\bm{\Theta})$ also provides a closed form of mean, variance and co-variance of each label's composition ratio in describing the conditioning sample $\bm{x}$. 
About this, we have the following theorem:
\begin{theorem}
Assuming the label distribution vector $\bm{\ell}$ follows the SNEFY conditional distribution $\mathrm{P}(d\bm{\ell}|\bm{x};\bm{V},\bm{\Theta})$ in Eq.~\eqref{eq:snefy_inner} with the kernel function $\bm{k}_{\sigma,\bm{t}_1,\bm{t}_2,\mu_1}(\bm{\theta}_i,\bm{\theta}_j;\bm{x})$ given in Eq.~\eqref{eq:kernel_dirichlet}, under the setting that $\bm{t}_1(\bm{\ell})=(\log\bm{\ell}^{y_1},\log\bm{\ell}^{y_2},\cdots,\log\bm{\ell}^{y_L})$, $\sigma=\exp$, and $\mu_1(d\bm{\ell})=d\bm{\ell}$, as well as the constraint that $\bm{W}_1>-1/2$ elementwise, for the $r$th label's composition ratio, $\bm{\ell}^{y_r}$, we have its conditional mean $\mathrm{E}[\bm{\ell}^{y_r}|\bm{x}]$ as
\begin{equation}
\mathrm{E}[\bm{\ell}^{y_r}|\bm{x}]=\frac{\mathrm{vec}(\bm{V}^{\top}\bm{V})^{\top}\mathrm{vec}(\bm{K}_{\bm{\Theta}}(\bm{x})\circ\bm{F}^{y_r})}{\mathrm{vec}(\bm{V}^{\top}\bm{V})^{\top}\mathrm{vec}(\bm{K}_{\bm{\Theta}}(\bm{x}))},
\label{eq:mean}
\end{equation}
%\textcolor{red}{Double check against above size of $\bm{V}$}
where $\circ$ denotes Hadamard product, and $\bm{F}^{y_r}$ is a $n\times n$ matrix, whose $ij$th entry is 
\begin{equation}
F_{ij}^{y_r}=\frac{1+w_{1ir}+w_{1jr}}{L+\sum_{l=1}^L(w_{1il}+w_{1jl})}.
\end{equation}
The conditional variance of $\bm{\ell}^{y_r}$, $\mathrm{Var}[\bm{\ell}^{y_r}|\bm{x}]$, is
\begin{equation}
%\begin{aligned}
\mathrm{Var}[\bm{\ell}^{y_r}|\bm{x}]=\frac{\mathrm{vec}(\bm{V}^{\top}\bm{V})^{\top}\mathrm{vec}(\bm{K}_{\bm{\Theta}}(\bm{x})\circ\bm{G}^{y_r})}{\mathrm{vec}(\bm{V}^{\top}\bm{V})^{\top}\mathrm{vec}(\bm{K}_{\bm{\Theta}}(\bm{x}))}\\
-\mathrm{E}^2[\bm{\ell}^{y_r}|\bm{x}],
%\end{aligned}
\label{eq:var}
\end{equation}
where $\bm{G}^{y_r}$ is a $n\times n$ matrix, with its $ij$th element being
\begin{equation}
%\begin{aligned}
G^{y_r}_{ij}=
\frac{(1+w_{1ir}+w_{1jr})(2+w_{1ir}+w_{1jr})}{[L+\sum_{l=1}^L(w_{1il}+w_{1jl})][1+L+\sum_{l=1}^L(w_{1il}+w_{1jl})]}.
%\end{aligned}
\end{equation}
For two different labels $y_r$ and $y_s$, with $y_r\neq y_s$, the conditional covariance of $\bm{\ell}^{y_r}$ and $\bm{\ell}^{y_s}$, $\mathrm{Cov}[\bm{\ell}^{y_r},\bm{\ell}^{y_s}|\bm{x}]$, is
\begin{equation}
%\begin{aligned}
\mathrm{Cov}[\bm{\ell}^{y_r},\bm{\ell}^{y_s}|\bm{x}]=\frac{\mathrm{vec}(\bm{V}^{\top}\bm{V})^{\top}\mathrm{vec}(\bm{K}_{\bm{\Theta}}(\bm{x})\circ\bm{H}^{y_r,y_s})}{\mathrm{vec}(\bm{V}^{\top}\bm{V})^{\top}\mathrm{vec}(\bm{K}_{\bm{\Theta}}(\bm{x}))}
-\mathrm{E}[\bm{\ell}^{y_r}|\bm{x}]\cdot\mathrm{E}[\bm{\ell}^{y_s}|\bm{x}],
%\end{aligned}
\end{equation}
where $\bm{H}^{y_r,y_s}$ is a $n\times n$ matrix 
%\textcolor{red}{Double check against above size of $\bm{V}$}
, with its $ij$th element being
\begin{equation}
%\begin{aligned}
H^{y_r,y_s}_{ij}=
\frac{(1+w_{1ir}+w_{1jr})(1+w_{1is}+w_{1js})}{[L+\sum_{l=1}^L(w_{1il}+w_{1jl})][1+L+\sum_{l=1}^L(w_{1il}+w_{1jl})]}.
%\end{aligned}
\end{equation}
\end{theorem}The proof is provided in the Appendix. 
Given the fitted distribution $\mathrm{P}(d\bm{\ell}|\bm{x};\bm{V},\bm{\Theta})$, the mean $\mathrm{E}[\bm{\ell}^{y_r}|\bm{x}]$ can be used to predict the unknown label distribution as the expectation over all values in the simplex. 
We can use the variance $\mathrm{Var}[\bm{\ell}^{y_r}|\bm{x}]$ to quantify label distribution prediction uncertainties. 
We can also use $\mathrm{E}[\bm{\ell}^{y_r}|\bm{x}]$ with $\mathrm{Var}[\bm{\ell}^{y_r}|\bm{x}]$ to construct confidence intervals for label distribution predictions by applying Chebyshev's inequality~\citep{grimmett2020probability}.
The covariance $\mathrm{Cov}[\bm{\ell}^{y_r},\bm{\ell}^{y_s}|\bm{x}]$ is helpful for us to understand the correlations between two different labels. 
More importantly, all the statistics are conditioned on the given sample $\bm{x}$, guiding us to make instance-wise decisions.

The conditional distribution $\mathrm{P}(d\bm{\ell}|\bm{x};\bm{V},\bm{\Theta})$ relies on the parameters $\bm{V}$ and $\bm{\Theta}$, as well as the neural network parameters for constructing $\bm{t}_2$ (we also use $\bm{t}_2$ to denote the parameters without confusion). 
We train the model with maximum likelihood estimation (MLE), by minimizing the negative log conditional  likelihoods on training samples:
\begin{equation}
\min_{\bm{V},\bm{\Theta},\bm{t}_2}-\sum_{\bm{x}^{\prime}\in\mathcal{X}}\log\frac{\mathrm{P}(d\bm{\ell}|\bm{x};\bm{V},\bm{\Theta})}{d\bm{\ell}}\bigg|_{\bm{x}=\bm{x}^{\prime},\bm{\ell}=\bm{\ell}_{\bm{x}^{\prime}}}.
\label{eq:erm}
\end{equation}
There are numerous metrics to measure the consistency between two label distributions, like Chebyshev distance, Kullback-Leibler divergence and Cosine coefficient~\citep{geng2016label}. 
Instead of optimizing these metrics, the MLE based learning objective in Eq.~\eqref{eq:erm} provides an alternative way to train the LDL model.
%The MLE based learning objective in Eq.~\eqref{eq:erm} can free us from the choice phobia about what metrics should be used to design LDL objectives. 
The fitted distribution $\mathrm{P}(d\bm{\ell}|\bm{x};\bm{V},\bm{\Theta})$ can be applied to various downstream tasks for quantifying the uncertainty of label distribution predictions.
%, and its closed-form mean is able to offer accurate label distribution predictions that are generic to varying evaluation metrics.

\textbf{Algorithm Description and Time Complexity.} We train the SNEFY-LDL model with stochastic gradient descent. 
The training procedure is shown in Algorithm 1. 
The model parameters $\bm{V}$, $\bm{\Theta}$ and $\bm{t}_2$ are first initialized by random numbers. 
We then iteratively select a batch of training samples, calculate the batched  likelihoods with Eq.~\eqref{eq:snefy_inner}, and  update parameters $\bm{V}$, $\bm{\Theta}$ and $\bm{t}_2$ by descending them along the gradient of batched negative log likelihoods. Taking the epoch number as a constant and assuming the latent layers of $\bm{t}_2$ have neurons in the same scale as the neuron number in the last layer $D_2$, the time complexity of Algorithm 1 is $\mathcal{O}(N(mn^2+Ln^2+dD_2+D_2^2))$, 
%\textcolor{red}{Check $m$ versus $n$}
which is linear to the number of training samples $N$, 
%and the input feature dimension $d$, 
making the algorithm able to scale to large datasets. 
For any sample $\bm{x}$ with an unknown label distribution, its label distribution mean and variance can be computed in time complexity $\mathcal{O}(mn^2+Ln^2+dD_2+D_2^2)$ using the closed-form formulations in Eq.~\eqref{eq:mean} and Eq.~\eqref{eq:var}.

\begin{algorithm}[tb]
	\caption{Training SNEFY-LDL}
	\label{alg:algorithm}
	\textbf{Input}: Training set $\{(\bm{x}_1,\bm{\ell}_{\bm{x}_1}),(\bm{x}_2,\bm{\ell}_{\bm{x}_2}),\cdots,(\bm{x}_N,\bm{\ell}_{\bm{x}_N})\}$.\\
	\textbf{Parameter}: $(\bm{V}, \bm{\Theta}, \bm{t}_2)$.\\
	\textbf{Output}: Optimized $(\bm{V}, \bm{\Theta}, \bm{t}_2)$.
	\begin{algorithmic}[1] %[1] enables line numbers
		\STATE $(\bm{V}, \bm{\Theta}, \bm{t}_2)$ $\leftarrow$ random initialization;
		\WHILE{epoch number does not expire}
		\STATE $\mathcal{B}\leftarrow$ randomly split training set into batches;
		\FOR{each batch in $\mathcal{B}$}
		\STATE Calculate batched $\bm{K}_{\bm{\Theta}}(\bm{x})$ with Eq.~\eqref{eq:kernel_dirichlet};
		\STATE Calculate batched likelihoods with Eq.~\eqref{eq:snefy_inner};
		\STATE $(\bm{V}, \bm{\Theta}, \bm{t}_2)$ $\leftarrow$ update by descending along the gradient of batched negative log likelihoods;
		\ENDFOR
		\ENDWHILE
		\STATE \textbf{return} optimized $(\bm{V}, \bm{\Theta}, \bm{t}_2)$.
	\end{algorithmic}
\end{algorithm}

\begin{table}[t]
	\centering
	\begin{tabular}{cccc}
		\toprule
		Dataset & \#Examples & \#Features & \#Labels\\
		\midrule
		Movie & 7,755 & 1,869 & 5 \\
		Natural Scene & 2,000 & 294 & 9 \\
		SBU\_3DFE & 2,500 & 243 & 6 \\
		SJAFFE & 213 & 243 & 6 \\
		\bottomrule
	\end{tabular}
	\caption{Summary of the four benchmark datasets.}
	\label{dataset}
\end{table}

\section{Experiments}
We conduct extensive experiments on conformal prediction, active learning and ensemble learning to verify SNEFY-LDL's ability in LDL uncertainty quantification. 

\renewcommand{\arraystretch}{0.95} 
\begin{table*}[t]
\centering
\tabcolsep=0.15cm
\resizebox{\textwidth}{!}{
\begin{tabular}{ccc|cc|cc}
\toprule
\multirow{2}{*}{Class Id} & \multicolumn{2}{c|}{bin size $=2$} & \multicolumn{2}{c|}{bin size $=4$} & \multicolumn{2}{c}{bin size $=8$}\\
& Dirichlet & SNEFY-LDL & Dirichlet & SNEFY-LDL & Dirichlet & SNEFY-LDL\\
\midrule
{\small 1} & {\small 0.8577$\pm$0.0310} & {\small \textbf{0.8747$\pm$0.0243}} & {\small 0.6987$\pm$0.2685} & {\small \textbf{0.8524$\pm$0.0298}} & {\small 0.6476$\pm$0.2505} & {\small \textbf{0.8335$\pm$0.0318}}\\
{\small 2} & {\small 0.8512$\pm$0.0333} & {\small \textbf{0.8867$\pm$0.0300}} & {\small 0.4878$\pm$0.3210} & {\small \textbf{0.8723$\pm$0.0361}} & {\small 0.4581$\pm$0.2952} & {\small \textbf{0.8277$\pm$0.0472}}\\
{\small 3} & {\small 0.8921$\pm$0.0208} & {\small \textbf{0.8924$\pm$0.0183}} & {\small 0.1644$\pm$0.3396} & {\small \textbf{0.8104$\pm$0.0576}} & {\small 0.1395$\pm$0.2904} & {\small \textbf{0.7537$\pm$0.0788}}\\
{\small 4} & {\small \textbf{0.8964$\pm$0.0176}} & {\small 0.8804$\pm$0.0256} & {\small 0.5403$\pm$0.3666} & {\small \textbf{0.8801$\pm$0.0257}} & {\small 0.4897$\pm$0.3246} & {\small \textbf{0.8188$\pm$0.0412}}\\
{\small 5} & {\small 0.8492$\pm$0.0300} & {\small \textbf{0.8690$\pm$0.0344}} & {\small \textbf{0.7561$\pm$0.2065}} & {\small 0.5276$\pm$0.3452} & {\small \textbf{0.6676$\pm$0.1905}} & {\small 0.4883$\pm$0.3184}\\
{\small 6} & {\small 0.8806$\pm$0.0248} & {\small \textbf{0.8890$\pm$0.0230}} & {\small 0.2005$\pm$0.3537} & {\small \textbf{0.8323$\pm$0.1615}} & {\small 0.1817$\pm$0.3193} & {\small \textbf{0.8149$\pm$0.1582}}\\
{\small 7} & {\small 0.8447$\pm$0.0335} & {\small \textbf{0.8662$\pm$0.0303}} & {\small 0.5131$\pm$0.3417} & {\small \textbf{0.8499$\pm$0.0340}} & {\small 0.4827$\pm$0.3210} & {\small \textbf{0.7697$\pm$0.0754}}\\
{\small 8} & {\small 0.8427$\pm$0.0329} & {\small \textbf{0.8897$\pm$0.0218}} & {\small 0.5107$\pm$0.3399} & {\small \textbf{0.8893$\pm$0.0219}} & {\small 0.4840$\pm$0.3175} & {\small \textbf{0.8753$\pm$0.0245}}\\
{\small 9} & {\small \textbf{0.8107$\pm$0.0548}} & {\small 0.7681$\pm$0.0472} & {\small 0.1443$\pm$0.2992} & {\small \textbf{0.4513$\pm$0.2982}} & {\small 0.1242$\pm$0.2609} & {\small \textbf{0.4096$\pm$0.2741}}\\
\bottomrule
\end{tabular}}
\caption{The conformal prediction performance measured by the FSC metric on the \textit{Natural\_Scene} dataset.}
\label{Conformal_natural}
\end{table*}

\begin{table*}[h]
\centering
\tabcolsep=0.15cm
\resizebox{\textwidth}{!}{
\begin{tabular}{ccc|cc|cc}
\toprule
\multirow{2}{*}{Class Id} & \multicolumn{2}{c|}{bin size $=2$} & \multicolumn{2}{c|}{bin size $=4$} & \multicolumn{2}{c}{bin size $=8$}\\
& Dirichlet & SNEFY-LDL & Dirichlet & SNEFY-LDL & Dirichlet & SNEFY-LDL\\
\midrule
{\small 1} & {\small 0.8913$\pm$0.0205} & {\small \textbf{0.8948$\pm$0.0175}} & {\small 0.8407$\pm$0.0511} & {\small \textbf{0.8836$\pm$0.0189}} & {\small 0.5391$\pm$0.3022} & {\small \textbf{0.8277$\pm$0.0675}}\\
{\small 2} & {\small \textbf{0.8816$\pm$0.0248}} & {\small 0.8658$\pm$0.0297} & {\small \textbf{0.8703$\pm$0.0266}} & {\small 0.8472$\pm$0.0364} & {\small \textbf{0.8475$\pm$0.0326}} & {\small 0.8210$\pm$0.0435}\\
{\small 3} & {\small 0.8737$\pm$0.0278} & {\small \textbf{0.8963$\pm$0.0175}} & {\small 0.8552$\pm$0.0292} & {\small \textbf{0.8785$\pm$0.0258}} & {\small 0.8290$\pm$0.0349} & {\small \textbf{0.8505$\pm$0.0297}}\\
{\small 4} & {\small 0.8848$\pm$0.0238} & {\small \textbf{0.8879$\pm$0.0222}} & {\small 0.8274$\pm$0.0367} & {\small \textbf{0.8725$\pm$0.0270}} & {\small 0.7193$\pm$0.1287} & {\small \textbf{0.7261$\pm$0.1500}}\\
{\small 5} & {\small 0.8937$\pm$0.0177} & {\small \textbf{0.8955$\pm$0.0184}} & {\small 0.8150$\pm$0.0462} & {\small \textbf{0.8381$\pm$0.0403}} & {\small 0.0793$\pm$0.2388} & {\small \textbf{0.4199$\pm$0.3376}}\\
{\small 6} & {\small 0.8911$\pm$0.0191} & {\small \textbf{0.8920$\pm$0.0205}} & {\small \textbf{0.8625$\pm$0.0366}} & {\small 0.8334$\pm$0.0453} & {\small \textbf{0.6025$\pm$0.2969}} & {\small 0.5048$\pm$0.3136}\\
\bottomrule
\end{tabular}}
\caption{The conformal prediction performance measured by the FSC metric on the \textit{SBU\_3DFE} dataset.}
\label{Conformal_sbu}
\end{table*}

\textbf{Benchmark Datasets.} We use four datasets~\citep{geng2016label} to benchmark our experiments, including the \textit{Movie} dataset containing label distributions on five movie rating scales, the \textit{Natural Scene} dataset with label distributions constructed by inconsistent multi-label ranking on natural scene images, the facial expression datasets \textit{SBU\_3DFE} and \textit{SJAFFE} with label distributions on six emotions. The statistics of the four benchmark datasets are summarized in Table \ref{dataset}. 

\textbf{Implementation Details.} When implementing SNEFY-LDL, $n$ and $m$ are respectively set to 64 and 32, $D_2$ is set as equal to $n$ and a one-layer neural network with ReLU activation is used to construct $\bm{t}_2$. 
The model is trained for 100 epochs with batch size 64 for conformal prediction and batch size 16 for active learning and ensemble learning. Weight clipping~\citep{arjovsky2017wasserstein} is used to control $\bm{W}_1>-1/2$ after each parameter update.

\subsection{Conformal Prediction}
As shown in \textbf{Theorem 3}, with the trained SNEFY-LDL model, given a new sample $\bm{x}$, we can get the closed-form conditional mean for the $r$th label's composition ratio as $\mathrm{E}[\bm{\ell}^{y_r}|\bm{x}]$ in Eq. (\ref{eq:mean}) and the closed-form conditional variance $\mathrm{Var}[\bm{\ell}^{y_r}|\bm{x}]$ in Eq. (\ref{eq:var}). By applying the Chebyshev's inequality~\citep{grimmett2020probability}, we can construct a confidence interval for the $r$th label's composition ratio in describing $\bm{x}$, $\bm{\ell}^{y_r}_{\bm{x}}$,
%given the conditional sample $\bm{x}$, 
which can be formally stated as
\begin{equation}
\mathrm{P}(|\bm{\ell}^{y_r}_{\bm{x}}-\mathrm{E}[\bm{\ell}^{y_r}|\bm{x}]|\leq k\sqrt{\mathrm{Var}[\bm{\ell}^{y_r}|\bm{x}]})\geq 1-\frac{1}{k^2},
\end{equation}i.e., the confidence interval of $\bm{\ell}^{y_r}_{\bm{x}}$ at the $1-1/k^2$ level is 
\begin{equation*}
\left[\mathrm{E}[\bm{\ell}^{y_r}|\bm{x}]-k\sqrt{\mathrm{Var}[\bm{\ell}^{y_r}|\bm{x}]}, \mathrm{E}[\bm{\ell}^{y_r}|\bm{x}]+k\sqrt{\mathrm{Var}[\bm{\ell}^{y_r}|\bm{x}]}\right],
\end{equation*}
which can be further calibrated as a conformal prediction task for the one-dimensional uncertainty estimate $\bm{\ell}^{y_r}_{\bm{x}}$. According to~\citet{angelopoulos2021gentle}, on a calibration set with $N_\text{cal}$ samples, we can define a calibration score function as 
\begin{equation}
s(\bm{x},\bm{\ell}^{y_r}_{\bm{x}})=\frac{\lvert\bm{\ell}^{y_r}_{\bm{x}}-\mathrm{E}[\bm{\ell}^{y_r}|\bm{x}]\rvert}{k\sqrt{\mathrm{Var}[\bm{\ell}^{y_r}|\bm{x}]}},
\label{eq:calibration_score}
\end{equation}where $\bm{\ell}^{y_r}_{\bm{x}}$ is the ground-truth value of the $r$th label's composition ratio in describing the calibration sample $\bm{x}$. By computing the calibration scores of all calibration samples, we can get the $\lceil(1-1/k^2)(N_\text{cal}+1)\rceil/N_\text{cal}$ quantile of the calibration scores as $\hat{q}_{y_r}$. For a new sample $\bm{x}$, the calibrated confidence interval for $\bm{\ell}^{y_r}_{\bm{x}}$ at the $1-1/k^2$ level is 
%$[\mathrm{E}[\bm{\ell}^{y_r}|\bm{x}]-k\cdot\hat{q}_{y_r}\sqrt{\mathrm{Var}[\bm{\ell}^{y_r}|\bm{x}]}, \mathrm{E}[\bm{\ell}^{y_r}|\bm{x}]+k\cdot\hat{q}_{y_r}\sqrt{\mathrm{Var}[\bm{\ell}^{y_r}|\bm{x}]}]$.
\begin{equation}
%\begin{aligned}
\mathcal{C}(\bm{x},\bm{\ell}^{y_r})=\left[\mathrm{E}[\bm{\ell}^{y_r}|\bm{x}]-k\cdot\hat{q}_{y_r}\sqrt{\mathrm{Var}[\bm{\ell}^{y_r}|\bm{x}]},
\mathrm{E}[\bm{\ell}^{y_r}|\bm{x}]+k\cdot\hat{q}_{y_r}\sqrt{\mathrm{Var}[\bm{\ell}^{y_r}|\bm{x}]}\right].
%\end{aligned}
\label{eq:confi_interval}
\end{equation}

\begin{table*}[t]
\centering
\tabcolsep=0.15cm
\resizebox{\textwidth}{!}{
\begin{tabular}{ccccccc}
\toprule
Method & Cheby $\downarrow$  & Clark $\downarrow$ & Canb $\downarrow$ & KL $\downarrow$ & Cos $\uparrow$ & Inter $\uparrow$ \\
\midrule
Random & {\small 0.1490$\pm$0.0098} & {\small 0.6359$\pm$0.0247} & {\small 1.1999$\pm$0.0448} & {\small 0.1409$\pm$0.0136} & {\small 0.9054$\pm$0.0093} & {\small 0.7950$\pm$0.0089} \\
Kmeans & {\small 0.1456$\pm$0.0076} & {\small 0.6242$\pm$0.0176} & {\small 1.1815$\pm$0.0320} & {\small 0.1361$\pm$0.0072} & {\small 0.9079$\pm$0.0068} & {\small 0.7981$\pm$0.0077} \\
CoreSet & {\small 0.1484$\pm$0.0164} & {\small 0.6322$\pm$0.0383} & {\small 1.1962$\pm$0.0720} & {\small 0.1399$\pm$0.0179} & {\small 0.9042$\pm$0.0158} & {\small 0.7941$\pm$0.0171} \\
Graph Density & {\small 0.1428$\pm$0.0059} & {\small 0.6175$\pm$0.0197} & {\small 1.1671$\pm$0.0346} & {\small 0.1326$\pm$0.0068} & {\small 0.9108$\pm$0.0057} & {\small 0.8013$\pm$0.0076} \\
Dirichlet & {\small 0.1472$\pm$0.0090} & {\small 0.6282$\pm$0.0245} & {\small 1.1899$\pm$0.0470} & {\small 0.1376$\pm$0.0116} & {\small 0.9064$\pm$0.0085} & {\small 0.7959$\pm$0.0099} \\
\midrule
SNEFY-LDL & {\small \textbf{0.1350$\pm$0.0030}} & {\small \textbf{0.5981$\pm$0.0134}} & {\small \textbf{1.1283$\pm$0.0254}} & {\small \textbf{0.1209$\pm$0.0049}} & {\small \textbf{0.9191$\pm$0.0030}} & {\small \textbf{0.8098$\pm$0.0042}} \\
\bottomrule
\end{tabular}}
\caption{The label distribution active learning performance on the \textit{Motive} dataset.}
\label{Motive_active}
\end{table*}

\begin{table*}[t]
\centering
\tabcolsep=0.15cm
\resizebox{\textwidth}{!}{
\begin{tabular}{ccccccc}
\toprule
Method & Cheby $\downarrow$  & Clark $\downarrow$ & Canb $\downarrow$ & KL $\downarrow$ & Cos $\uparrow$ & Inter $\uparrow$ \\
\midrule
Random & {\small 0.3591$\pm$0.0197} & {\small 2.4871$\pm$0.0280} & {\small 6.9131$\pm$0.1456} & {\small 0.9835$\pm$0.0766} & {\small 0.6554$\pm$0.0411} & {\small 0.4528$\pm$0.0336}\\
Kmeans & {\small 0.3690$\pm$0.0163} & {\small 2.4929$\pm$0.0200} & {\small 6.9407$\pm$0.0808} & {\small 1.0498$\pm$ 0.1078} & {\small 0.6318$\pm$0.0375} & {\small 0.4361$\pm$0.0228}\\
CoreSet & {\small 0.3629$\pm$0.0193} & {\small 2.4842$\pm$0.0233} & {\small 6.8882$\pm$0.1089} & {\small 1.0102$\pm$0.1078} & {\small 0.6442$\pm$0.0489} & {\small 0.4485$\pm$0.0340}\\
Graph Density & {\small 0.3700$\pm$0.0242} & {\small 2.4958$\pm$0.0277} & {\small 6.9643$\pm$0.1394} & {\small 1.0471$\pm$0.0991} & {\small 0.6230$\pm$0.0453} & {\small 0.4294$\pm$0.0321} \\
Dirichlet & {\small 0.3720$\pm$0.0213} & {\small 2.4951$\pm$0.0226} & {\small 6.9502$\pm$0.0977}   & {\small 1.0628$\pm$0.0909} & {\small 0.6249$\pm$0.0336} & {\small 0.4326$\pm$0.0192} \\
\midrule
SNEFY-LDL & {\small \textbf{0.3474$\pm$0.0160}} & {\small \textbf{2.4807$\pm$0.0194}} & {\small \textbf{6.8755$\pm$0.0845}} & {\small \textbf{0.9244$\pm$0.0648}} & {\small \textbf{0.6819$\pm$0.0285}}   & {\small \textbf{0.4718$\pm$0.0203}} \\
\bottomrule
\end{tabular}}
\caption{The label distribution active learning performance on the \textit{Natural\_Scene} dataset.}
\label{Natural_Scene_active}
\end{table*}

As a baseline, we extend the competitive SA-BFGS~\citep{geng2016label} algorithm by modeling the distribution of (rather than point-estimating) the label distribution vectors.
We model the distribution using the Dirichlet distribution centered at the point prediction.
%As a baseline, we model the distribution of label distribution vectors using the Dirichlet distribution centered at the label distribution predictions yielded by the competitive SA-BFGS~\citep{geng2016label} algorithm. 
Following the same routine of conformal prediction, we can also construct calibrated confidence intervals for label affiliation probabilities given new samples. Following~\citet{angelopoulos2021gentle}, we use the Feature-Stratified Coverage (FSC) metric to evaluate the adaptivity of the constructed confidence intervals in Eq.(\ref{eq:confi_interval}), which categorizes the test samples into different groups by dividing the numeric values at the first feature dimension into different bins, then compute the coverage rate of the confidence intervals in each group, and picks up the lowest group-level coverage rate as the final metric value. We select the \textit{Natural\_Scene} and \textit{SBU\_3DFE} datasets with a fair number of examples and labels to test the performance of conformal prediction. We randomly split each dataset into the training, calibration and test sets according to the ratio of 50\%/25\%/25\% for 100 times and report the average FSC scores with bin size equal to 2, 4 and 8. In the experiment, we aim to construct 90\% level confidence intervals by setting $1/k^2=0.1$, which means that FSC scores closer to 0.9 indicates better conformal prediction performance. 

Tables~\ref{Conformal_natural}-\ref{Conformal_sbu} compare the conformal prediction performance measured by the FSC metric on the \textit{Natural\_Scene} and \textit{SBU\_3DFE} datasets. For each comparison between SNEFY-LDL and Dirichlet, the better performer is highlighted by \textbf{boldface}. From the tables, we can find that the proposed SNEFY-LDL model outperforms the Dirichlet baseline in most cases. SNEFY-LDL constructs a multimodal distribution to model the distribution of label distribution vectors on the probability simplex space, which is more flexible than the unimodal Dirichlet, and contributes to the label distribution confidence intervals with greater adaptivity. 

\subsection{Active Learning}

To further evaluate SNEFY-LDL's performance in uncertainty quantification, we choose the \textit{Movie} and \textit{Natural\_Scene} datasets to conduct the active learning experiments. We first randomly split the two datasets into the training and test sets according to the ratio of 90\%/10\% for ten times. For each training-test set split, we randomly select 400 labeled training samples to form the initial labeled pool and take the remaining training samples as the unlabeled pool. We first train a SNEFY-LDL model with the labeled samples in the initial labeled pool. We then use different active learning strategies to pick up 100 informative samples from the unlabeled pool and query their labels. After augmenting the initial labeled pool with the 100 queried samples, we re-train the SNEFY-LDL model and evaluate the performance of its label distribution predictions produced by the closed-form conditional mean in Eq. (\ref{eq:mean}). Following \citet{geng2016label}, the label distribution prediction performance is evaluated by the following six metrics: Chebyshev distance (Cheby), Clark distance (Clark), Canberra metric (Canb), Kullback-Leibler divergence (KL), Cosine coefficient (Cos) and Intersection (Inter). 
The average scores on the ten random training-test splits are reported. Six different active learning strategies are compared:

\begin{itemize}
	\item \textbf{Random}~\citep{zhan2022comparative} randomly selects 100 samples from the unlabeled pool.
	\item \textbf{Kmeans}~\citep{zhdanov2019diverse} selects samples close to the cluster centroids generated by the Kmeans clustering~\citep{mackay2003information} in feature space. 
	\item \textbf{CoreSet}~\citep{sener2018active} selects the $k$-center samples~\citep{har2011geometric} as representative unlabeled samples, which is a variant of Kmeans.  
	\item \textbf{Graph Density}~\citep{ebert2012ralf} selects highly connected samples in the constructed KNN graph~\citep{preparata2012computational}.
	\item \textbf{Dirichlet} models the distribution of label distributions using a Dirichlet distribution~\citep{ng2011dirichlet} centered at predicted label distributions and selects samples with the largest differential entropy scores~\citep{cover1999elements}. 
	\item \textbf{SNEFY-LDL} uses importance sampling~\citep{kloek1978bayesian} to estimate the differential entropy values of the conditional distributions modeled by SNEFY-LDL and picks up samples that have the largest differential entropy scores.
\end{itemize}

\begin{table*}[t]
\centering
\tabcolsep=0.1cm
\resizebox{\textwidth}{!}{
\begin{tabular}{cccccccc}
\toprule
Base Learner & Bagging & Cheby $\downarrow$  & Clark $\downarrow$ & Canb $\downarrow$ & KL $\downarrow$ & Cos $\uparrow$ & Inter $\uparrow$ \\
\midrule
\multirow{2}{*}{SA-BFGS}& {\small Average} & {\small 0.1178$\pm$0.0020} & {\small 0.3743$\pm$0.0068} & {\small 0.7948$\pm$0.0163} & {\small 0.0641$\pm$0.0022} & {\small 0.9370$\pm$0.0020} & {\small 0.8575$\pm$0.0028}\\
& {\small Weighted} & {\small \textbf{0.1137$\pm$0.0023}} & {\small \textbf{0.3625$\pm$0.0062}} & {\small \textbf{0.7686$\pm$0.0149}} & {\small \textbf{0.0604$\pm$0.0021}} & {\small \textbf{0.9406$\pm$0.0020}} & {\small \textbf{0.8624$\pm$0.0026}} \\
\midrule
\multirow{2}{*}{DF-LDL}& {\small Average} & {\small 0.1203$\pm$0.0017} & {\small 0.3762$\pm$0.0059} & {\small 0.8040$\pm$0.0146} & {\small 0.0657$\pm$0.0019} & {\small 0.9353$\pm$0.0018} & {\small 0.8557$\pm$0.0025}\\
& {\small Weighted} & {\small \textbf{0.1152$\pm$0.0024}} & {\small \textbf{0.3617$\pm$0.0070}} & {\small \textbf{0.7715$\pm$0.0168}} & {\small \textbf{0.0609$\pm$0.0024}} & {\small \textbf{0.9399$\pm$0.0023}} & {\small \textbf{0.8617$\pm$0.0030}} \\
\midrule
\multirow{2}{*}{LDL-SCL}& {\small Average} & {\small 0.1256$\pm$0.0020} & {\small 0.3828$\pm$0.0047} & {\small 0.8260$\pm$0.0118} & {\small 0.0699$\pm$0.0021} & {\small 0.9315$\pm$0.0018} & {\small 0.8519$\pm$0.0020}\\
& {\small Weighted} & {\small \textbf{0.1246$\pm$0.0023}} & {\small \textbf{0.3772$\pm$0.0048}} & {\small \textbf{0.8147$\pm$0.0114}} & {\small \textbf{0.0684$\pm$0.0022}} & {\small \textbf{0.9330$\pm$0.0019}} & {\small \textbf{0.8540$\pm$0.0020}} \\
\midrule
\multirow{2}{*}{LDL-LRR}& {\small Average} & {\small 0.1269$\pm$0.0021} & {\small 0.3966$\pm$0.0052} & {\small 0.8478$\pm$0.0131} & {\small 0.0730$\pm$0.0022} & {\small 0.9285$\pm$0.0019} & {\small 0.8478$\pm$0.0023}\\
& {\small Weighted} & {\small \textbf{0.1250$\pm$0.0020}} & {\small \textbf{0.3916$\pm$0.0044}} & {\small \textbf{0.8373$\pm$0.0106}} & {\small \textbf{0.0710$\pm$0.0020}} & {\small \textbf{0.9305$\pm$0.0018}} & {\small \textbf{0.8498$\pm$0.0019}} \\
\bottomrule
\end{tabular}}
\caption{The label distribution ensemble learning performance on the \textit{SBU\_3DFE} dataset.}
\label{SBU_3DFE_ensemble}
\end{table*}

\begin{table*}[t]
\centering
\tabcolsep=0.1cm
\resizebox{\textwidth}{!}{
\begin{tabular}{cccccccc}
\toprule
Base Learner & Bagging & Cheby $\downarrow$  & Clarky $\downarrow$ & Canb $\downarrow$ & KL $\downarrow$ & Cos $\uparrow$ & Inter $\uparrow$ \\
\midrule
\multirow{2}{*}{SA-BFGS}& {\small Average} & {\small 0.0889$\pm$0.0085} & {\small 0.3180$\pm$0.0197} & {\small 0.6529$\pm$0.0461} & {\small 0.0406$\pm$0.0058} & {\small 0.9613$\pm$0.0057} & {\small 0.8890$\pm$0.0086}\\
& {\small Weighted} & {\small \textbf{0.0842$\pm$0.0084}} & {\small \textbf{0.3118$\pm$0.0200}} & {\small \textbf{0.6390$\pm$0.0466}} & {\small \textbf{0.0385$\pm$0.0056}} & {\small \textbf{0.9636$\pm$0.0056}} & {\small \textbf{0.8923$\pm$0.0089}}\\
\midrule
\multirow{2}{*}{DF-LDL}& {\small Average} & {\small 0.0951$\pm$0.0093} & {\small 0.3385$\pm$0.0253} & {\small 0.6958$\pm$0.0611} & {\small 0.0456$\pm$0.0078} & {\small 0.9566$\pm$0.0072} & {\small 0.8818$\pm$0.0110}\\
& {\small Weighted} & {\small \textbf{0.0881$\pm$0.0092}} & {\small \textbf{0.3189$\pm$0.0258}} & {\small \textbf{0.6525$\pm$0.0617}} & {\small \textbf{0.0408$\pm$0.0074}} & {\small \textbf{0.9612$\pm$0.0069}} & {\small
\textbf{0.8895$\pm$0.0110}}\\
\midrule
\multirow{2}{*}{LDL-SCL}& {\small Average} & {\small 0.0911$\pm$0.0090} & {\small 0.3249$\pm$0.0219} & {\small 0.6746$\pm$0.0493} & {\small 0.0424$\pm$0.0063} & {\small 0.9596$\pm$0.0062} & {\small 0.8854$\pm$0.0092}\\
& {\small Weighted} & {\small \textbf{0.0865$\pm$0.0095}} & {\small \textbf{0.3153$\pm$0.0231}} & {\small \textbf{0.6509$\pm$0.0540}} & {\small \textbf{0.0400$\pm$0.0064}} & {\small \textbf{0.9621$\pm$0.0064}} & {\small \textbf{0.8898$\pm$0.0100}}\\
\midrule
\multirow{2}{*}{LDL-LRR}& {\small Average} & {\small 0.0888$\pm$0.0090} & {\small 0.3189$\pm$0.0209} & {\small 0.6538$\pm$0.0473} & {\small 0.0408$\pm$0.0065} & {\small 0.9612$\pm$0.0063} & {\small 0.8890$\pm$0.0090}\\
& {\small Weighted} & {\small \textbf{0.0846$\pm$0.0089}} & {\small \textbf{0.3124$\pm$0.0197}} & {\small \textbf{0.6392$\pm$0.0444}} & {\small \textbf{0.0387$\pm$0.0062}} & {\small \textbf{0.9634$\pm$0.0063}} & {\small \textbf{0.8922$\pm$0.0087}}\\
\bottomrule
\end{tabular}}
\caption{The label distribution ensemble learning performance on the \textit{SJAFFE} dataset.}
\label{SJAFFE_ensemble}
\end{table*}

%Among the six active learning strategies, Kmeans, CoreSet and Graph Density are representativeness based methods, which rely on only sample features by characterizing samples' geometric properties in feature space, while Dirichlet and SNEFY-LDL are the uncertainty based methods we have contrived, which leverage the predictions of the initially trained SNEFY-LDL model. 

Tables~\ref{Motive_active}-\ref{Natural_Scene_active} compare the performance of different active learning strategies, where the best performer is highlighted by \textbf{boldface}. As is shown in the tables, SNEFY-LDL consistently achieves the best performance in terms of all metrics. By accurately evaluating label distribution prediction uncertainties, SNEFY-LDL can pick up more informative unlabeled samples than the naive uncertainty quantification strategy, Dirichlet, as well as the representativeness based active learning strategies, Kmeans, CoreSet and Graph Density, which are even sometimes inferior to the Random strategy.

\subsection{Ensemble Learning}
We also conduct experiments on ensemble learning to further verify SNEFY-LDL's ability in uncertainty quantification, with the expectation that reliable base learners can be identified by the SNEFY-LDL probability modeling. Bagging~\citep{breiman1996bagging} is adopted as an exemplary ensemble learning paradigm. We choose the \textit{SEU\_3DFE} and \textit{SJAFFE} datasets, and randomly split them into training and test sets according to the ratio of 90\%/10\%. For each training-test set split, we randomly select 50 samples from the training set for 25 rounds, train 25 base LDL learners with the selected samples, and evaluate the label distribution prediction performance of the ensembled LDL model on the test set. Four competitive LDL models are employed as base learners: SA-BFGS~\citep{geng2016label}, DF-LDL~\citep{gonzalez2021decomposition}, LDL-SCL~\citep{jia2019label} and LDL-LRR~\citep{jia2023adaptive}, and two strategies are adopted to ensemble base learner predictions: 1) \textbf{Average}: aggregate the 25 base learner predictions with the uniform weight $1/25$, and 2) \textbf{Weighted}: weight each base learner prediction in proportion to its corresponding SNEFY-LDL probability density conditioned on each test sample in an instance-wise manner. 

Tables~\ref{SBU_3DFE_ensemble}-\ref{SJAFFE_ensemble} compare the two different ensemble learning strategies, where the best strategy is highlighted by \textbf{boldface}. 
From the tables, we find that the \textbf{Weighted} strategy is significantly better than \textbf{Average} in terms of all metrics. 
This implies that SNEFY-LDL provides an effective mechanism to quantify the reliability of base learners' label distribution predictions so that the reliable base learners are highlighted to contribute to a better ensemble learning performance. 

\section{Conclusion}

%In this paper, 
We propose a novel LDL paradigm: estimate the distribution of label distribution vectors on the probability simplex,
%to quantify the uncertainties of label distribution predictions.
%conditioned on the input sample, 
which brings a bird's-eye view on the relative significance of all possible label distributions. 
By uncovering the underlying relationship between SNEFY and LDL, we develop the SNEFY-LDL model that can provide a tractable formulation of the conditional distribution of label distribution vectors, enjoying great expressivity and high computational efficiency. 
SNEFY-LDL admits closed-form expressions for the distribution's mean, variance and covariance, 
%that can be computed quickly,
%in constant time
making SNEFY-LDL able to provide real-time responses in real-world applications. 
Experiments on conformal prediction, active learning and ensemble learning demonstrate the great utility of SNEFY-LDL for uncertainty-aware applications. 

\bibliography{uai2025-SNEFY-LDL}

%\newpage

\onecolumn

%\title{Label Distribution Learning \\using the Squared Neural Family on the Probability Simplex\\(Supplementary Material)}
%\maketitle

%This Supplementary Material should be submitted together with the main paper.
\appendix
\section{Theorem Proofs}
\begin{customthm}{2}
Let $\bm{t}_1(\bm{\ell})=(\log\bm{\ell}^{y_1},\log\bm{\ell}^{y_2},\cdots,\log\bm{\ell}^{y_L}) : \Delta^{L-1}\rightarrow\mathbb{R}^{L}$ by setting $D_1=L$, the activation function $\sigma$ be the exponential function $\exp$, the base measure $\mu_1(d\bm{\ell})=d\bm{\ell}$ be the Lebesgue measure. Under the condition that $\bm{W}_1>-1/2$ elementwise, the kernel function $\bm{k}_{\sigma,\bm{t}_1,\bm{t}_2,\mu_1}(\bm{\theta}_i,\bm{\theta}_j;\bm{x})$ admits a closed form:
\begin{equation}
\bm{k}_{\bm{t}_2}(\bm{\theta}_i,\bm{\theta}_j;\bm{x})=\exp(\bm{w}_{2i}^{\top}\bm{t}_2(\bm{x})+\bm{w}_{2j}^{\top}\bm{t}_2(\bm{x})+b_i+b_j)\cdot
\frac{\prod_{l=1}^{L}\Gamma(1+w_{1il}+w_{1jl})}{\Gamma\big(L+\sum_{l=1}^{L}(w_{1il}+w_{1jl})\big)},
\tag{9}
%\label{eq:kernel_dirichlet}
\end{equation}
where $w_{1il}$ is the $il$-th element of matrix $\bm{W}_1$ and $\Gamma(\cdot)$ is the gamma function.
\end{customthm}
\begin{proof}
According to Eq.~\eqref{eq:condition_kernel}, 
\begin{equation*}
\begin{aligned}
\bm{k}_{\sigma,\bm{t}_1,\bm{t}_2,\mu_1}(\bm{\theta}_i,\bm{\theta}_j;\bm{x})=&\int_{\Delta^{L-1}}\tilde{\bm{k}}_{\sigma,\bm{t}_1,\bm{t}_2}(\bm{\theta}_i,\bm{\theta}_j;\bm{\ell},\bm{x})\mu_1(d\bm{\ell})\\
=&\int_{\Delta^{L-1}}\sigma(\bm{w}_{1i}^{\top}\bm{t}_1(\bm{\ell})+\bm{w}_{2i}^{\top}\bm{t}_2(\bm{x})+b_i)\cdot\sigma(\bm{w}_{1j}^{\top}\bm{t}_1(\bm{\ell})+\bm{w}_{2j}^{\top}\bm{t}_2(\bm{x})+b_j)\mu_1(d\bm{\ell}).
\end{aligned}
\end{equation*}
Given the setting $\bm{t}_1(\bm{\ell})=(\log\bm{\ell}^{y_1},\log\bm{\ell}^{y_2},\cdots,\log\bm{\ell}^{y_L})$, $\sigma=\exp$ and $\mu_1(d\bm{\ell})=d\bm{\ell}$, $\bm{k}_{\sigma,\bm{t}_1,\bm{t}_2,\mu_1}$ can be written as
\begin{equation*}
\begin{aligned}
\bm{k}_{\bm{t}_2}(\bm{\theta}_i,\bm{\theta}_j;\bm{x})&=\int_{\Delta^{L-1}}\exp(\bm{w}_{1i}^{\top}\bm{t}_1(\bm{\ell})+\bm{w}_{2i}^{\top}\bm{t}_2(\bm{x})+b_i)\cdot\exp(\bm{w}_{1j}^{\top}\bm{t}_1(\bm{\ell})+\bm{w}_{2j}^{\top}\bm{t}_2(\bm{x})+b_j)d\bm{\ell}\\
&=\exp(\bm{w}_{2i}^{\top}\bm{t}_2(\bm{x})+\bm{w}_{2j}^{\top}\bm{t}_2(\bm{x})+b_i+b_j)\cdot\int_{\Delta^{L-1}}\prod_{l=1}^{L}(\bm{\ell}^{y_l})^{w_{1il}+w_{1jl}}d\bm{\ell}.\\
\end{aligned}
\end{equation*}
As $\bm{W}_{1}>-1/2$ elementwise, $w_{1il}+w_{1jl}+1>0$. Assuming $\bm{\ell}$ follows a Dirichlet distribution 
%(Ng, Tian, and Tang 2011) 
with parameters $\bm{\alpha}=(\alpha_1,\alpha_2,\cdots,\alpha_L)$, where $\alpha_l=w_{1il}+w_{1jl}+1>0$, its probability density, $\mathrm{P}_{\mathrm{Dir}}(d\bm{\ell})/{d\bm{\ell}}$, is in the form:
\begin{equation*}
\frac{\mathrm{P}_{\mathrm{Dir}}(d\bm{\ell})}{d\bm{\ell}}=\frac{1}{\mathrm{B}(\bm{\alpha})}\prod_{l=1}^{L}(\bm{\ell}^{y_l})^{\alpha_l-1},
\end{equation*}where $\mathrm{B}(\cdot)$ is the beta function. 
Considering the fact that $\int_{\Delta^{L-1}}\mathrm{P}_{\mathrm{Dir}}(d\bm{\ell})=1$, 
\begin{equation*}
\int_{\Delta^{L-1}}\prod_{l=1}^{L}(\bm{\ell}^{y_l})^{\alpha_l-1}d\bm{\ell}=\mathrm{B}(\bm{\alpha}). 
\end{equation*}
That is to say
\begin{equation*}
\int_{\Delta^{L-1}}\prod_{l=1}^{L}(\bm{\ell}^{y_l})^{w_{1il}+w_{1jl}}d\bm{\ell}=\frac{\prod_{l=1}^{L}\Gamma(1+w_{1il}+w_{1jl})}{\Gamma\big(L+\sum_{l=1}^{L}(w_{1il}+w_{1jl})\big)}. 
\end{equation*}
Therefore,
\begin{equation*}
\bm{k}_{\bm{t}_2}(\bm{\theta}_i,\bm{\theta}_j;\bm{x})=\exp(\bm{w}_{2i}^{\top}\bm{t}_2(\bm{x})+\bm{w}_{2j}^{\top}\bm{t}_2(\bm{x})+b_i+b_j)\cdot
\frac{\prod_{l=1}^{L}\Gamma(1+w_{1il}+w_{1jl})}{\Gamma\big(L+\sum_{l=1}^{L}(w_{1il}+w_{1jl})\big)}.
\end{equation*}
\end{proof}

\begin{customthm}{3}
Assuming the label distribution vector $\bm{\ell}$ follows the SNEFY conditional distribution $\mathrm{P}(d\bm{\ell}|\bm{x};\bm{V},\bm{\Theta})$ in Eq.~\eqref{eq:snefy_inner} with the kernel function $\bm{k}_{\sigma,\bm{t}_1,\bm{t}_2,\mu_1}(\bm{\theta}_i,\bm{\theta}_j;\bm{x})$ given in Eq.~\eqref{eq:kernel_dirichlet}, under the setting that $\bm{t}_1(\bm{\ell})=(\log\bm{\ell}^{y_1},\log\bm{\ell}^{y_2},\cdots,\log\bm{\ell}^{y_L})$, $\sigma=\exp$, and $\mu_1(d\bm{\ell})=d\bm{\ell}$, as well as the constraint that $\bm{W}_1>-1/2$ elementwise, for the $r$th label's composition ratio, $\bm{\ell}^{y_r}$, we have its conditional mean $\mathrm{E}[\bm{\ell}^{y_r}|\bm{x}]$ as
\begin{equation}
\mathrm{E}[\bm{\ell}^{y_r}|\bm{x}]=\frac{\mathrm{vec}(\bm{V}^{\top}\bm{V})^{\top}\mathrm{vec}(\bm{K}_{\bm{\Theta}}(\bm{x})\circ\bm{F}^{y_r})}{\mathrm{vec}(\bm{V}^{\top}\bm{V})^{\top}\mathrm{vec}(\bm{K}_{\bm{\Theta}}(\bm{x}))},
\tag{10}
%\label{eq:mean}
\end{equation}
%\textcolor{red}{Double check against above size of $\bm{V}$}
where $\circ$ denotes Hadamard product, and $\bm{F}^{y_r}$ is a $n\times n$ matrix, whose $ij$th entry is 
\begin{equation}
F_{ij}^{y_r}=\frac{1+w_{1ir}+w_{1jr}}{L+\sum_{l=1}^L(w_{1il}+w_{1jl})}.
\tag{11}
\end{equation}
The conditional variance of $\bm{\ell}^{y_r}$, $\mathrm{Var}[\bm{\ell}^{y_r}|\bm{x}]$, is
\begin{equation}
\mathrm{Var}[\bm{\ell}^{y_r}|\bm{x}]=\frac{\mathrm{vec}(\bm{V}^{\top}\bm{V})^{\top}\mathrm{vec}(\bm{K}_{\bm{\Theta}}(\bm{x})\circ\bm{G}^{y_r})}{\mathrm{vec}(\bm{V}^{\top}\bm{V})^{\top}\mathrm{vec}(\bm{K}_{\bm{\Theta}}(\bm{x}))}-\mathrm{E}^2[\bm{\ell}^{y_r}|\bm{x}],
\tag{12}
\end{equation}
where $\bm{G}^{y_r}$ is a $n\times n$ matrix 
%\textcolor{red}{Double check against above size of $\bm{V}$}
, with its $ij$th element being
\begin{equation}
G^{y_r}_{ij}=
\frac{(1+w_{1ir}+w_{1jr})(2+w_{1ir}+w_{1jr})}{[L+\sum_{l=1}^L(w_{1il}+w_{1jl})][1+L+\sum_{l=1}^L(w_{1il}+w_{1jl})]}.
\tag{13}
\end{equation}
For two different labels $y_r$ and $y_s$, with $y_r\neq y_s$, the conditional covariance of $\bm{\ell}^{y_r}$ and $\bm{\ell}^{y_s}$, $\mathrm{Cov}[\bm{\ell}^{y_r},\bm{\ell}^{y_s}|\bm{x}]$, is
\begin{equation}
\mathrm{Cov}[\bm{\ell}^{y_r},\bm{\ell}^{y_s}|\bm{x}]=\frac{\mathrm{vec}(\bm{V}^{\top}\bm{V})^{\top}\mathrm{vec}(\bm{K}_{\bm{\Theta}}(\bm{x})\circ\bm{H}^{y_r,y_s})}{\mathrm{vec}(\bm{V}^{\top}\bm{V})^{\top}\mathrm{vec}(\bm{K}_{\bm{\Theta}}(\bm{x}))}-\mathrm{E}[\bm{\ell}^{y_r}|\bm{x}]\cdot\mathrm{E}[\bm{\ell}^{y_s}|\bm{x}],
\tag{14}
\end{equation}
where $\bm{H}^{y_r,y_s}$ is a $n\times n$ matrix 
%\textcolor{red}{Double check against above size of $\bm{V}$}
, with its $ij$th element being
\begin{equation}
H^{y_r,y_s}_{ij}=
\frac{(1+w_{1ir}+w_{1jr})(1+w_{1is}+w_{1js})}{[L+\sum_{l=1}^L(w_{1il}+w_{1jl})][1+L+\sum_{l=1}^L(w_{1il}+w_{1jl})]}.
\tag{15}
\end{equation}
\end{customthm}
\begin{proof}
For any fixed scalar function $\varphi(\bm{\ell})$ of $\bm{\ell}$ with $\varphi(\cdot):\Delta^{L-1}\rightarrow \mathbb{R}$, its
expectation with regard to the conditional SNEFY distribution in Eq.~\eqref{eq:snefy_inner}, $\mathrm{E}[\varphi(\bm{\ell})|\bm{x}]$, can be computed as
\begin{equation*}
\begin{aligned}
\mathrm{E}[\varphi(\bm{\ell})|\bm{x}]=&\int_{\Delta^{L-1}}\varphi(\bm{\ell})\mathrm{P}(d\bm{\ell}|\bm{x};\bm{V},\bm{\Theta})\\
=&\int_{\Delta^{L-1}}\varphi(\bm{\ell})\frac{\mathrm{vec}(\bm{V}^{\top}\bm{V})^{\top}\mathrm{vec}(\widetilde{\bm{K}}_{\bm{\Theta}}(\bm{\ell},\bm{x}))}{\mathrm{vec}(\bm{V}^{\top}\bm{V})^{\top}\mathrm{vec}(\bm{K}_{\bm{\Theta}}(\bm{x}))}\mu_1(d\bm{\ell})\\=
&\frac{\mathrm{vec}(\bm{V}^{\top}\bm{V})^{\top}\mathrm{vec}(\bm{\Phi}_{\bm{\Theta}}(\bm{x}))}{\mathrm{vec}(\bm{V}^{\top}\bm{V})^{\top}\mathrm{vec}(\bm{K}_{\bm{\Theta}}(\bm{x}))},
\end{aligned}
\end{equation*}where $\bm{\Phi}_{\bm{\Theta}}(\bm{x})\in\mathbb{R}^{n\times n}$ is the elementwise integral:
\begin{equation*}
\bm{\Phi}_{\bm{\Theta}}(\bm{x})=\int_{\Delta^{L-1}}\varphi(\bm{\ell})\widetilde{\bm{K}}_{\bm{\Theta}}(\bm{\ell},\bm{x})\mu_1(d\bm{\ell}),
\end{equation*}whose $ij$th element is
\begin{equation*}
\begin{aligned}
&\bm{\phi}_{\sigma,\bm{t}_1,\bm{t}_2,\mu_1}(\bm{\theta}_i,\bm{\theta}_j;\bm{x})=\int_{\Delta^{L-1}}\varphi(\bm{\ell})\tilde{\bm{k}}_{\sigma,\bm{t}_1,\bm{t}_2}(\bm{\theta}_i,\bm{\theta}_j;\bm{\ell},\bm{x})\mu_1(d\bm{\ell})\\
&=\int_{\Delta^{L-1}}\varphi(\bm{\ell})\sigma(\bm{w}_{1i}^{\top}\bm{t}_1(\bm{\ell})+\bm{w}_{2i}^{\top}\bm{t}_2(\bm{x})+b_i)\cdot\sigma(\bm{w}_{1j}^{\top}\bm{t}_1(\bm{\ell})+\bm{w}_{2j}^{\top}\bm{t}_2(\bm{x})+b_j)\mu_1(d\bm{\ell}).\\
\end{aligned}
\end{equation*}By setting $\bm{t}_1(\bm{\ell})=(\log\bm{\ell}^{y_1},\log\bm{\ell}^{y_2},\cdots,\log\bm{\ell}^{y_L})$, $\sigma=\exp$, and $\mu_1(d\bm{\ell})=d\bm{\ell}$, $\bm{\phi}_{\sigma,\bm{t}_1,\bm{t}_2,\mu_1}(\bm{\theta}_i,\bm{\theta}_j;\bm{x})$ can be written as
\begin{equation*}
\begin{aligned}
\bm{\phi}_{\bm{t}_2}(\bm{\theta}_i,\bm{\theta}_j;\bm{x})&=\int_{\Delta^{L-1}}\varphi(\bm{\ell})\exp(\bm{w}_{1i}^{\top}\bm{t}_1(\bm{\ell})+\bm{w}_{2i}^{\top}\bm{t}_2(\bm{x})+b_i)\cdot\exp(\bm{w}_{1j}^{\top}\bm{t}_1(\bm{\ell})+\bm{w}_{2j}^{\top}\bm{t}_2(\bm{x})+b_j)d\bm{\ell}\\
&=\exp(\bm{w}_{2i}^{\top}\bm{t}_2(\bm{x})+\bm{w}_{2j}^{\top}\bm{t}_2(\bm{x})+b_i+b_j)\cdot\int_{\Delta^{L-1}}\varphi(\bm{\ell})\prod_{l=1}^{L}(\bm{\ell}^{y_l})^{w_{1il}+w_{1jl}}d\bm{\ell}.\\
\end{aligned}
\end{equation*}
As $\bm{W}_{1}>-1/2$ elementwise, $w_{1il}+w_{1jl}+1>0$. Assuming $\bm{\ell}$ follows a Dirichlet distribution with parameters $\bm{\alpha}=(\alpha_1,\alpha_2,\cdots,\alpha_L)$, where $\alpha_l=w_{1il}+w_{1jl}+1>0$, its probability density, $\mathrm{P}_{\mathrm{Dir}}(d\bm{\ell})/d\bm{\ell}$, is in the form:
\begin{equation*}
\frac{\mathrm{P}_{\mathrm{Dir}}(d\bm{\ell})}{d\bm{\ell}}=\frac{1}{\mathrm{B}(\bm{\alpha})}\prod_{l=1}^{L}(\bm{\ell}^{y_l})^{\alpha_l-1},
\end{equation*}where $\mathrm{B}(\cdot)$ is the beta function. Then, we have
\begin{equation*}
\begin{aligned}
\int_{\Delta^{L-1}}\varphi(\bm{\ell})\prod_{l=1}^{L}(\bm{\ell}^{y_l})^{w_{1il}+w_{1jl}}d\bm{\ell}&=\mathrm{B}(\bm{\alpha})\int_{{\Delta}^{L-1}}\varphi(\bm{\ell})\mathrm{P}_{\mathrm{Dir}}(d\bm{\ell})\\
&=\mathrm{B}(\bm{\alpha})\mathrm{E}_{\mathrm{Dir}}[\varphi(\bm{\ell});\bm{w}_{1i},\bm{w}_{1j}]\\
&=\frac{\prod_{l=1}^{L}\Gamma(1+w_{1il}+w_{1jl})}{\Gamma\big(L+\sum_{l=1}^{L}(w_{1il}+w_{1jl})\big)}\mathrm{E}_{\mathrm{Dir}}[\varphi(\bm{\ell});\bm{w}_{1i},\bm{w}_{1j}],\\
\end{aligned}
\end{equation*}where $\mathrm{E}_{\mathrm{Dir}}[\varphi(\bm{\ell});\bm{w}_{1i},\bm{w}_{1j}]$ is the expectation of $\varphi(\bm{\ell})$ with regard to the Dirichlet distribution parameterized by $\bm{w}_{1i}$ and $\bm{w}_{1j}$. Therefore, 
\begin{equation*}
\begin{aligned}
\bm{\phi}_{\bm{t}_2}(\bm{\theta}_i,\bm{\theta}_j;\bm{x})
=&\exp(\bm{w}_{2i}^{\top}\bm{t}_2(\bm{x})+\bm{w}_{2j}^{\top}\bm{t}_2(\bm{x})+b_i+b_j)
\cdot\frac{\prod_{l=1}^{L}\Gamma(1+w_{1il}+w_{1jl})}{\Gamma\big(L+\sum_{l=1}^{L}(w_{1il}+w_{1jl})\big)}\mathrm{E}_{\mathrm{Dir}}[\varphi(\bm{\ell});\bm{w}_{1i},\bm{w}_{1j}]\\
=&\bm{k}_{\bm{t}_2}(\bm{\theta}_i,\bm{\theta}_j;\bm{x})\mathrm{E}_{\mathrm{Dir}}[\varphi(\bm{\ell});\bm{w}_{1i},\bm{w}_{1j}].
\end{aligned}
\end{equation*}
By using $\bm{E}_{\varphi}$ to denote the $n\times n$ matrix whose $ij$th element is $\mathrm{E}_{\mathrm{Dir}}[\varphi(\bm{\ell});\bm{w}_{1i},\bm{w}_{1j}]$, we have
\begin{equation*}
\bm{\Phi}_{\bm{\Theta}}(\bm{x})=\bm{K}_{\bm{\Theta}}(\bm{x})\circ\bm{E}_{\varphi},
\end{equation*}
\begin{equation*}
\mathrm{E}[\varphi(\bm{\ell})|\bm{x}]=\frac{\mathrm{vec}(\bm{V}^{\top}\bm{V})^{\top}\mathrm{vec}(\bm{K}_{\bm{\Theta}}(\bm{x})\circ\bm{E}_{\varphi})}{\mathrm{vec}(\bm{V}^{\top}\bm{V})^{\top}\mathrm{vec}(\bm{K}_{\bm{\Theta}}(\bm{x}))}.
\end{equation*}
For the Dirichlet distribution, $\mathrm{E}_{\mathrm{Dir}}[\varphi(\bm{\ell});\bm{w}_{1i},\bm{w}_{1j}]$ has closed forms for some moments.  In particular, for $\varphi(\bm{\ell})=\bm{\ell}^{y_r}$, $\varphi(\bm{\ell})=(\bm{\ell}^{y_r})^2$, and $\varphi(\bm{\ell})=\bm{\ell}^{y_r}\cdot\bm{\ell}^{y_s}$ with $y_r\neq y_s$, we respectively have 
\begin{equation*}
\mathrm{E}_{\mathrm{Dir}}[\bm{\ell}^{y_r};\bm{w}_{1i},\bm{w}_{1j}]=\frac{1+w_{1ir}+w_{1jr}}{L+\sum_{l=1}^L(w_{1il}+w_{1jl})},
\end{equation*}
\begin{equation*}
\mathrm{E}_{\mathrm{Dir}}[(\bm{\ell}^{y_r})^2;\bm{w}_{1i},\bm{w}_{1j}]=
\frac{(1+w_{1ir}+w_{1jr})(2+w_{1ir}+w_{1jr})}{[L+\sum_{l=1}^L(w_{1il}+w_{1jl})][1+L+\sum_{l=1}^L(w_{1il}+w_{1jl})]},
\end{equation*}
\begin{equation*}
\mathrm{E}_{\mathrm{Dir}}[\bm{\ell}^{y_r}\cdot\bm{\ell}^{y_s};\bm{w}_{1i},\bm{w}_{1j}]=
\frac{(1+w_{1ir}+w_{1jr})(1+w_{1is}+w_{1js})}{[L+\sum_{l=1}^L(w_{1il}+w_{1jl})][1+L+\sum_{l=1}^L(w_{1il}+w_{1jl})]}.
\end{equation*}
Finally, we have
\begin{equation*}
\begin{aligned}
&\mathrm{E}[\bm{\ell}^{y_r}|\bm{x}]=\frac{\mathrm{vec}(\bm{V}^{\top}\bm{V})^{\top}\mathrm{vec}(\bm{K}_{\bm{\Theta}}(\bm{x})\circ\bm{F}^{y_r})}{\mathrm{vec}(\bm{V}^{\top}\bm{V})^{\top}\mathrm{vec}(\bm{K}_{\bm{\Theta}}(\bm{x}))},\\
&\mathrm{E}[(\bm{\ell}^{y_r})^2|\bm{x}]=\frac{\mathrm{vec}(\bm{V}^{\top}\bm{V})^{\top}\mathrm{vec}(\bm{K}_{\bm{\Theta}}(\bm{x})\circ\bm{G}^{y_r})}{\mathrm{vec}(\bm{V}^{\top}\bm{V})^{\top}\mathrm{vec}(\bm{K}_{\bm{\Theta}}(\bm{x}))},\\
&\mathrm{E}[\bm{\ell}^{y_r}\cdot\bm{\ell}^{y_s}|\bm{x}]=\frac{\mathrm{vec}(\bm{V}^{\top}\bm{V})^{\top}\mathrm{vec}(\bm{K}_{\bm{\Theta}}(\bm{x})\circ\bm{H}^{y_r,y_s})}{\mathrm{vec}(\bm{V}^{\top}\bm{V})^{\top}\mathrm{vec}(\bm{K}_{\bm{\Theta}}(\bm{x}))},\\
\end{aligned}
\end{equation*}where $\bm{F}^{y_r}$, $\bm{G}^{y_r}$ and $\bm{H}^{y_r,y_s}$ denote the $n\times n$ matrices whose $ij$th elements are $\mathrm{E}_{\mathrm{Dir}}[\bm{\ell}^{y_r};\bm{w}_{1i},\bm{w}_{1j}]$, $\mathrm{E}_{\mathrm{Dir}}[(\bm{\ell}^{y_r})^2;\bm{w}_{1i},\bm{w}_{1j}]$ and $\mathrm{E}_{\mathrm{Dir}}[\bm{\ell}^{y_r}\cdot\bm{\ell}^{y_s};\bm{w}_{1i},\bm{w}_{1j}]$ respectively. 

$\mathrm{Var}[\bm{\ell}^{y_r}|\bm{x}]$ and $\mathrm{Cov}[\bm{\ell}^{y_r},\bm{\ell}^{y_s}|\bm{x}]$ can be directly derived by using the identities that $\mathrm{Var}[\bm{\ell}^{y_r}|\bm{x}]=\mathrm{E}[(\bm{\ell}^{y_r})^2|\bm{x}]-\mathrm{E}^2[\bm{\ell}^{y_r}|\bm{x}]$ and $\mathrm{Cov}[\bm{\ell}^{y_r},\bm{\ell}^{y_s}|\bm{x}]=\mathrm{E}[\bm{\ell}^{y_r}\cdot\bm{\ell}^{y_s}|\bm{x}]-\mathrm{E}[\bm{\ell}^{y_r}|\bm{x}]\cdot\mathrm{E}[\bm{\ell}^{y_s}|\bm{x}]$. 
\end{proof}

%\section{Additional simulation results}
%Table~\ref{tab:supp-data} lists additional simulation results; see also \citet{einstein} for a comparison. 

%\begin{table}[!h]
%    \centering
%    \caption{An Interesting Table.} \label{tab:supp-data}
%    \begin{tabular}{rl}
%        \toprule % from booktabs package
%        \bfseries Dataset & \bfseries Result\\
%        \midrule % from booktabs package
%        Data1 & 0.12345\\
%        Data2 & 0.67890\\
%        Data3 & 0.54321\\
%        Data4 & 0.09876\\
%        \bottomrule % from booktabs package
%    \end{tabular}
%\end{table}

%\section{Math font exposition}
% NOTE: necessary when ptmx or no mathfont class option is given
%\providecommand{\upGamma}{\Gamma}
%\providecommand{\uppi}{\pi}
%How math looks in equations is important:
%\begin{equation*}
%    F_{\alpha,\beta}^\eta(z) = \upGamma(\tfrac{3}{2}) \prod_{\ell=1}^\infty\eta \frac{z^\ell}{\ell} + \frac{1}{2\uppi}\int_{-\infty}^z\alpha \sum_{k=1}^\infty x^{\beta k}\mathrm{d}x.
%\end{equation*}
%However, one should not ignore how well math mixes with text:
%The frobble function \(f\) transforms zabbies \(z\) into yannies \(y\).
%It is a polynomial \(f(z)=\alpha z + \beta z^2\), where \(-n<\alpha<\beta/n\leq\gamma\), with \(\gamma\) a positive real number.

\section{Experimental Details}
\subsection{Conformal Prediction}
We select the \textit{Natural Scene} and \textit{SBU\_3DFE} datasets for the conformal prediction experiments, as they have a relatively large number of labels that give more freedom in the change of label distribution compositions to warrant more uncertainties in label distribution predictions, and they also have enough samples to do the training/calibration/test set split. According to the ratio
of 50\%/25\%/25\%, we randomly split each dataset into the training, calibration and test sets. Here, we denote the training, calibration and test sets as $\mathcal{S}^{\text{tr}}=\{(\bm{x}^{\text{tr}}_i,\bm{\ell}_{\bm{x}^{\text{tr}}_i}):i=1,\cdots,N_{\text{tr}}\}$, $\mathcal{S}^{\text{cal}}=\{(\bm{x}^{\text{cal}}_i,\bm{\ell}_{\bm{x}^{\text{cal}}_i}):i=1,\cdots,N_{\text{cal}}\}$ and $\mathcal{S}^{\text{te}}=\{(\bm{x}^{\text{te}}_i,\bm{\ell}_{\bm{x}^{\text{te}}_i}):i=1,\cdots,N_{\text{te}}\}$ respectively, with $N_{\text{tr}}$, $N_{\text{cal}}$ and $N_{\text{te}}$ denoting the number of samples in the training, calibration and test sets respectively. For each split, a SNEFY-LDL is first trained on the training set $\mathcal{S}^{\text{tr}}$. The trained SNEFY-LDL is then leveraged to construct a 90\% level confidence interval for each label's composition ratio with SNEFY-LDL's closed-form conditional mean and variance. The confidence intervals constructed by the trained SNEFY-LDL model are then calibrated with the calibration set. The adaptivity of the calibrated confidence intervals are finally evaluated on the test set, as an indicator of SNEFY-LDL's ability in quantifying the prediction uncertainty of each label's composition ratio. On the test set, the adaptivity of the calibrated confidence intervals $\mathcal{C}(\bm{x},\bm{\ell}^{y_r})$ in Eq. (\ref{eq:confi_interval}) for the $r$th label's composition ratio $\bm{\ell}^{y_r}$ is measured by the Feature-Stratified Coverage (FSC) metric~\citep{angelopoulos2021gentle}. For calculating the FSC metric, we first bin the first feature of $\bm{x}$ for all $\bm{x}\in\mathcal{S}_{\text{te}}$ into a number of categories, $1,\cdots,G$, then categorize test samples in $\mathcal{S}_{\text{te}}$ into different groups $\{\mathcal{S}^{\text{te}}_g:g=1,\cdots,G\}$ according to the first feature's category values. Here, $G$ is termed as bin size or group number. The FSC metric for the confidence intervals of the $r$th label's composition ratio $\bm{\ell}^{y_r}$  is evaluated as the minimal coverage rate among the groups $\{\mathcal{S}^{\text{te}}_g:g=1,\cdots,G\}$:
\begin{equation}
\text{FSC}(\bm{\ell}^{y_r}):=\min_{g\in\{1,\cdots,G\}}\frac{1}{|\mathcal{S}_{g}^{\text{te}}|}\sum_{\bm{x}\in\mathcal{S}_g^{\text{te}}}\mathbbm{1}\left\{\bm{\ell}_{\bm{x}}^{y_r}\in\mathcal{C}(\bm{x},\bm{\ell}^{y_r})\right\}.
\label{eq:FSC}
\end{equation}
The detailed procedure of conformal prediction with SNEFY-LDL is described by Algorithm \ref{alg:algorithm_A1}.
\setcounter{algorithm}{0}
\renewcommand{\thealgorithm}{A\arabic{algorithm}}
\begin{algorithm}[h]
\caption{Conformal Prediction with SNEFY-LDL}
\label{alg:algorithm_A1}
\textbf{Input}: Training set $\mathcal{S}^{\text{tr}}=\{(\bm{x}^{\text{tr}}_i,\bm{\ell}_{\bm{x}^{\text{tr}}_i}):i=1,\cdots,N_{\text{tr}}\}$, calibration set $\mathcal{S}^{\text{cal}}=\{(\bm{x}^{\text{cal}}_i,\bm{\ell}_{\bm{x}^{\text{cal}}_i}):i=1,\cdots,N_{\text{cal}}\}$ and test set $\mathcal{S}^{\text{te}}=\{(\bm{x}^{\text{te}}_i,\bm{\ell}_{\bm{x}^{\text{te}}_i}):i=1,\cdots,N_{\text{te}}\}$, as well as the given label set $\mathcal{Y}=\{y_1,y_2,\cdots,y_{L}\}$.\\
\textbf{Parameter}: $1/k^2=0.1$ for constructing 90\% level confidence intervals.\\
\textbf{Output}: FSC scores on the test set.
\begin{algorithmic}[1] %[1] enables line numbers
\STATE Train a SNEFY-LDL model on the training set $\mathcal{S}^{\text{tr}}$ with Algorithm \ref{alg:algorithm};
\FOR{each sample $\bm{x}\in\mathcal{S}^{\text{cal}}$}
\FOR{each label $y_r\in\mathcal{Y}$}
\STATE Calculate the calibration score $s(\bm{x},\bm{\ell}^{y_r}_{\bm{x}})$ with Eq.(\ref{eq:calibration_score}) and the trained SNEFY-LDL model;
\ENDFOR
\ENDFOR
\FOR{each label $y_r\in\mathcal{Y}$}
\STATE Calculate the $\lceil(1-1/k^2)(N_\text{cal}+1)\rceil/N_\text{cal}$ quantile as $\hat{q}_{y_r}$ among the $N_\text{cal}$ calibration scores $s(\bm{x},\bm{\ell}^{y_r}_{\bm{x}})$ with $\bm{x}\in\mathcal{S}^{\text{cal}}$; 
\ENDFOR
\FOR{each sample $\bm{x}\in\mathcal{S}^{\text{te}}$}
\FOR{each label $y_r\in\mathcal{Y}$}
\STATE Construct the calibrated confidence interval $\mathcal{C}(\bm{x},\bm{\ell}^{y_r})$ with Eq. (\ref{eq:confi_interval}) and the calculated quantile $\hat{q}_{y_r}$;
\ENDFOR
\ENDFOR
\FOR{each label $y_r\in\mathcal{Y}$}
\STATE Evaluate the $\text{FSC}(\bm{\ell}^{y_r})$ score with Eq. (\ref{eq:FSC}) and the $N_{\text{te}}$ confidence intervals $\mathcal{C}(\bm{x},\bm{\ell}^{y_r})$ with $\bm{x}\in\mathcal{S}^{\text{te}}$;
\ENDFOR
\STATE \textbf{return} the evaluated $\text{FSC}(\bm{\ell}^{y_r})$ scores for all $y_r\in\mathcal{Y}$.
\end{algorithmic}
\end{algorithm}

\subsection{Active Learning}
The \textit{Motive} and \textit{Natural Scene} datasets are selected for active learning, as they are relatively sensitive to the label sparsity issue, more suitable to benchmark the performance change with informative samples labeled and augmented to the training data. We randomly split the two datasets into the training and test sets according to the ratio of 90\%/10\% for ten times. We denote the training set as $\mathcal{S}^{\text{tr}}=\{(\bm{x}^{\text{tr}}_i,\bm{\ell}_{\bm{x}^{\text{tr}}_i}):i=1,\cdots,N_{\text{tr}}\}$ and test set as $\mathcal{S}^{\text{te}}=\{(\bm{x}^{\text{te}}_i,\bm{\ell}_{\bm{x}^{\text{te}}_i}):i=1,\cdots,N_{\text{te}}\}$, where $N_{\text{tr}}$ and $N_{\text{te}}$ are respectively the number of samples in the training and test sets. For each training-test set split, we randomly select 400 labeled samples from the training set $\mathcal{S}^{\text{tr}}$ to form the initial labeled pool and take the remaining samples in $\mathcal{S}^{\text{tr}}$ as unlabeled samples. We first train a SNEFY-LDL model with the initial labeled pool. To achieve active learning with the trained SNEFY-LDL model, we first evaluate the differential entropy $H(\bm{x};\bm{V},\bm{\Theta})$ for each unlabeled sample $\bm{x}$ as
\begin{equation}
H(\bm{x};\bm{V},\bm{\Theta})=-\int_{\Delta^{L-1}}\left\{\log\frac{\mathrm{P}(d\bm{\ell}|\bm{x};\bm{V},\bm{\Theta})}{d\bm{\ell}}\right\}\mathrm{P}(d \bm{\ell}|\bm{x} ; \bm{V}, \bm{\Theta}),
\label{eq:differential_entropy}
\end{equation}then select 100 most informative unlabeled samples with the largest differential entropy values. After querying the labels of the 100 selected samples, we augment them into the initial labeled pool, and re-train another SNEFY-LDL model with the augmented labeled pool. With the re-trained SNEFY-LDL model, we can predict the label distribution vectors for samples in the test set given their feature vectors $\bm{x}$ by directly using the closed-form conditional mean in Eq.(\ref{eq:mean}). As a criterion of active learning, the label distribution prediction performance on the test set is evaluated using the six metrics~\citep{geng2016label}: Chebyshev distance (Cheby), Clark distance (Clark), Canberra metric (Canb), Kullback-Leibler divergence (KL), Cosine coefficient (Cos) and Intersection (Inter). Given the ground-truth and predicted label distribution vectors as $\bm{\ell}$ and $\hat{\bm{\ell}}$ respectively, the evaluation metrics are defined as follows:
\begin{equation}
\begin{aligned}
&\text{Cheby}(\bm{\ell},\hat{\bm{\ell}})=\Vert \bm{\ell}-\hat{\bm{\ell}}\Vert_\infty \downarrow,\quad\text{Clark}(\bm{\ell},\hat{\bm{\ell}})=\Big\Vert\frac{\bm{\ell}-\hat{\bm{\ell}}}{\bm{\ell}+\hat{\bm{\ell}}} \Big\Vert_2 \downarrow,
\quad\text{Canb}(\bm{\ell},\hat{\bm{\ell}})=\Big\Vert\frac{\bm{\ell}-\hat{\bm{\ell}}}{\bm{\ell}+\hat{\bm{\ell}}} \Big\Vert_1 \downarrow,\\
&\text{KL}(\bm{\ell},\hat{\bm{\ell}})=\sum_{y\in\mathcal{Y}}\bm{\ell}^{y}\log\frac{\bm{\ell}^{y}}{\hat{\bm{\ell}}^{y}}\downarrow,\;\;\;\text{Cos}(\bm{\ell},\hat{\bm{\ell}})=\frac{\bm{\ell}^\top \hat{\bm{\ell}}}{\Vert \bm{\ell} \Vert_2 \Vert \hat{\bm{\ell}} \Vert_2} \uparrow,\quad\;\;\text{Inter}(\bm{\ell},\hat{\bm{\ell}})=\sum_{y\in\mathcal{Y}}\min(\bm{\ell}^{y},\hat{\bm{\ell}}^{y})\uparrow.
\end{aligned}
\label{eq:metric}
\end{equation} 
For each metric, $\uparrow$ ($\downarrow$) indicates that higher (lower) scores imply better label distribution prediction performance. 

However, the differential entropy $H(\bm{x};\bm{V},\bm{\Theta})$ in Eq. (\ref{eq:differential_entropy}) cannot be computed in a closed form. To overcome this difficulty, we adopt importance sampling~\citep{kloek1978bayesian} to approximately estimate the differential entropy values, where the uniform distribution is selected as the proposal distribution. The detailed procedure is provided in Algorithm \ref{alg:algorithm_A2}. 
\begin{algorithm}[h]
\caption{Differential Entropy Estimation with Importance Sampling}
\label{alg:algorithm_A2}
\textbf{Input}: The feature vector of an unlabeled sample $\bm{x}$ and the conditional distribution $\mathrm{P}(d\bm{\ell}|\bm{x};\bm{V},\bm{\Theta})$ modeled by the trained SNEFY-LDL model, and the label set size $L$.\\
\textbf{Parameter}: The number of sampling iterations $N_{\text{iter}}=1,000$.\\
\textbf{Output}: The estimated differential entropy for the unlabeled sample $\bm{x}$, $\hat{H}(\bm{x};\bm{V},\bm{\Theta})$.
\begin{algorithmic}[1] %[1] enables line numbers
\FOR{each iteration $i\in\{1,\cdots,N_{\text{iter}}\}$}
\STATE Sample a label distribution vector $\tilde{\bm{\ell}}^{(i)}$ from the uniform distribution over the simplex $\Delta^{L-1}$ with probability density $q(\tilde{\bm{\ell}}^{(i)})=(L-1)!$;
\STATE Evaluate the probability density value $\mathrm{P}(d\bm{\ell}|\bm{x};\bm{V},\bm{\Theta})/d\bm{\ell}$ at $\bm{\ell}=\tilde{\bm{\ell}}^{(i)}$ as $p(\tilde{\bm{\ell}}^{(i)}|\bm{x};\bm{V},\bm{\Theta})$ with Eq. (\ref{eq:snefy_inner}) and the trained SNEFY-LDL model;
\ENDFOR
\STATE Calculate the approximated differential entropy $\hat{H}(\bm{x};\bm{V},\bm{\Theta})=-
\frac{1}{N_{\text{iter}}}\sum_{i=1}^{N_\text{iter}}\frac{p(\tilde{\bm{\ell}}^{(i)}|\bm{x};\bm{V},\bm{\Theta})}{q(\tilde{\bm{\ell}}^{(i)})}\log p(\tilde{\bm{\ell}}^{(i)}|\bm{x};\bm{V},\bm{\Theta})$;
\STATE \textbf{return} the estimated differential entropy $\hat{H}(\bm{x};\bm{V},\bm{\Theta})$ for the given unlabeled sample $\bm{x}$.
\end{algorithmic}
\end{algorithm}

The detailed procedure for active learning with SNEFY-LDL is described by Algorithm \ref{alg:algorithm_A3}. 

\subsection{Ensemble Learning}
The \textit{SBU\_3DFE} and \textit{SJAFFE} datasets are selected for the ensemble learning experiments, as they have a relatively small number of features, more efficient to train a number of base learners and do ensemble prediction. We randomly split the two datasets into the training and test sets according to the ratio of 90\%/10\% for ten times. We denote the training set as $\mathcal{S}^{\text{tr}}=\{(\bm{x}^{\text{tr}}_i,\bm{\ell}_{\bm{x}^{\text{tr}}_i}):i=1,\cdots,N_{\text{tr}}\}$ and test set as $\mathcal{S}^{\text{te}}=\{(\bm{x}^{\text{te}}_i,\bm{\ell}_{\bm{x}^{\text{te}}_i}):i=1,\cdots,N_{\text{te}}\}$, where $N_{\text{tr}}$ and $N_{\text{te}}$ are respectively the number of samples in the training and test sets. For each training-test set split, we randomly select 50 samples from the training set for 25 rounds, train 25 base LDL learners with the selected samples, and evaluate the label distribution prediction performance of the ensembled LDL model on the test set. Four competitive LDL algorithms are employed to train base learners: SA-BFGS~\citep{geng2016label}, DF-LDL~\citep{gonzalez2021decomposition}, LDL-SCL~\citep{jia2019label} and LDL-LRR~\citep{jia2023adaptive}. To achieve ensemble learning with SNEFY-LDL, we first train a SNEFY-LDL model with the training set $\mathcal{S}^{\text{tr}}$, and then do the ensemble prediction by weighting the base learners according to the SNEFY-LDL conditional probability densities measured at their predictions given each test sample $\bm{x}\in\mathcal{S}^{\text{te}}$. The detailed procedure is shown in Algorithm \ref{alg:algorithm_A4}.

\begin{algorithm}[t]
\caption{Active Learning with SNEFY-LDL}
\label{alg:algorithm_A3}
\textbf{Input}: Training set $\mathcal{S}^{\text{tr}}=\{(\bm{x}^{\text{tr}}_i,\bm{\ell}_{\bm{x}^{\text{tr}}_i}):i=1,\cdots,N_{\text{tr}}\}$ and test set $\mathcal{S}^{\text{te}}=\{(\bm{x}^{\text{te}}_i,\bm{\ell}_{\bm{x}^{\text{te}}_i}):i=1,\cdots,N_{\text{te}}\}$.\\
\textbf{Parameters}: The size of the initial labeled pool $N_{\text{initial}}=400$ and the number of queried samples $N_{\text{query}}=100$.\\
\textbf{Output}: The label distribution prediction performance scores measured by Cheby, Clark, Canb, KL, Cos and Inter on the test set $\mathcal{S}^{\text{te}}$.
\begin{algorithmic}[1] %[1] enables line numbers
\STATE Randomly select $N_{\text{initial}}$ samples from the training set $\mathcal{S}^{\text{tr}}$ to form the labeled pool $\mathcal{S}^{\text{tr}}_{\text{label}}$ and use the remaining samples to form the unlabeled pool $\mathcal{S}^{\text{tr}}_{\text{unlabel}}$;
\STATE Train a SNEFY-LDL model with the initial labeled pool;
\FOR{each sample $\bm{x}\in\mathcal{S}^{\text{tr}}_{\text{unlabel}}$}
\STATE Estimate the differential entropy value $\hat{H}(\bm{x};\bm{V},\bm{\Theta})$ for $\bm{x}$ with Algorithm \ref{alg:algorithm_A2} and the trained SNEFY-LDL model;
\ENDFOR
\STATE Select $N_\text{query}$ samples from $\mathcal{S}^{\text{tr}}_{\text{unlabel}}$ with the top-$N_\text{query}$ differential entropy values and query their label distributions;
\STATE Augment the $N_\text{query}$ queried samples into the labeled pool $\mathcal{S}^{\text{tr}}_{\text{label}}$;
\STATE Re-train another SNEFY-LDL model with the augmented labeled pool $\mathcal{S}^{\text{tr}}_{\text{label}}$;
\STATE Evaluate the label distribution prediction performance of the re-trained SNEFY-LDL model on the test set $\mathcal{S}^{\text{te}}$ (as an average over all test samples) with the metrics of Cheby, Clark, Canb, KL, Cos and Inter defined in Eq. (\ref{eq:metric});
\STATE \textbf{return} the label distribution prediction performance scores measured by Cheby, Clark, Canb, KL, Cos and Inter.
\end{algorithmic}
\end{algorithm}

\begin{algorithm}[H]
\caption{Ensemble Learning with SNEFY-LDL}
\label{alg:algorithm_A4}
\textbf{Input}: Training set $\mathcal{S}^{\text{tr}}=\{(\bm{x}^{\text{tr}}_i,\bm{\ell}_{\bm{x}^{\text{tr}}_i}):i=1,\cdots,N_{\text{tr}}\}$, test set $\mathcal{S}^{\text{te}}=\{(\bm{x}^{\text{te}}_i,\bm{\ell}_{\bm{x}^{\text{te}}_i}):i=1,\cdots,N_{\text{te}}\}$, and a LDL base learner training algorithm $\in\{\text{SA-BFGS},\text{DF-LDL},\text{LDL-SCL},\text{LDL-LRR}\}$.\\
\textbf{Parameters}: The number of samples for training base learners $N_\text{sample}=50$ and the number of base learners $N_{\text{base}}=25$.\\
\textbf{Output}: The label distribution prediction performance scores measured by Cheby, Clark, Canb, KL, Cos and Inter on the test set $\mathcal{S}^{\text{te}}$.
\begin{algorithmic}[1] %[1] enables line numbers
\FOR{each iteration $i\in\{1,\cdots,N_{\text{base}}\}$}
\STATE Randomly select $N_\text{sample}$ samples from the training set $\mathcal{S}^{\text{tr}}$;
\STATE Train a LDL base learner $B_i$ with the selected samples;
\ENDFOR
\STATE Train a SNEFY-LDL model with the training set $\mathcal{S}^{\text{tr}}$;
\FOR{each sample $\bm{x}\in\mathcal{S}^{\text{te}}$}
\FOR{each iteration $i\in\{1,\cdots,N_{\text{base}}\}$}
\STATE Predict $\bm{x}$'s label distribution vector with base learner $B_i$ as $\tilde{\bm{\ell}}^{(i)}_{\bm{x}}$;
\STATE Evaluate the probability density value $\mathrm{P}(d\bm{\ell}|\bm{x};\bm{V},\bm{\Theta})/d\bm{\ell}$ at $\bm{\ell}=\tilde{\bm{\ell}}^{(i)}_{\bm{x}}$ as $p(\tilde{\bm{\ell}}^{(i)}_{\bm{x}}|\bm{x};\bm{V},\bm{\Theta})$ with Eq. (\ref{eq:snefy_inner}) and the trained SNEFY-LDL model;
\ENDFOR
\STATE Predict $\bm{x}$'s label distribution vector $\tilde{\bm{\ell}}_{\bm{x}}$ as the weighted average of $\tilde{\bm{\ell}}^{(i)}_{\bm{x}}$, i.e., $\tilde{\bm{\ell}}_{\bm{x}}=\frac{\sum_{i=1}^{N_\text{base}}p(\tilde{\bm{\ell}}^{(i)}_{\bm{x}}|\bm{x};\bm{V},\bm{\Theta})\tilde{\bm{\ell}}^{(i)}_{\bm{x}}}{\sum_{i=1}^{N_\text{base}}p(\tilde{\bm{\ell}}^{(i)}_{\bm{x}}|\bm{x};\bm{V},\bm{\Theta})}$;
\ENDFOR
\STATE Evaluate the performance of the ensembled label distribution predictions $\tilde{\bm{\ell}}^{(i)}_{\bm{x}}$ on the test set $\mathcal{S}^{\text{te}}$ (as an average over all test samples) with the metrics of Cheby, Clark, Canb, KL, Cos and Inter defined in Eq. (\ref{eq:metric});
\STATE \textbf{return} the label distribution prediction performance scores measured by Cheby, Clark, Canb, KL, Cos and Inter.
\end{algorithmic}
\end{algorithm}

\section{Parameter Sensitivity Study}
By choosing the \textit{Natural Scene} dataset and the conformal prediction task, we take turns to study the sensitivity of SNEFY-LDL with regard to the four hyperparameters: $n$ and $m$, as well as the batch size and epoch number used for training, by varying the studied hyperparameter in a predefined range and fixing the remaining three as default values at each turn. For the conformal prediction task, the default values of $n$, $m$, batch size and epoch number are 64, 32, 64 and 100 respectively.   
Figure~\ref{fig:para} plots the conformal prediction performance change of SNEFY-LDL measured by FSC with bin size equal to 2 and 4 when the values of hyperparameters vary in a range.
From Figure~\ref{fig:para}, we can find that the performance of SNEFY-LDL remains relatively stable with the change of the four hyperparameters in most cases, except for the cases with $m=48$ and $96$ in Figure A\ref{fig:para:subfig:m}, as well as batch size $=32$ and $96$ in Figure A\ref{fig:para:subfig:batch}, where the conformal prediction performance of SNEFY-LDL goes through a obvious drop.

\setcounter{figure}{0}
\renewcommand{\thefigure}{A\arabic{figure}}
\begin{figure*}[t]
\centering
\subfigure[$n$]{
\label{fig:para:subfig:n} 
\includegraphics[width=3in]{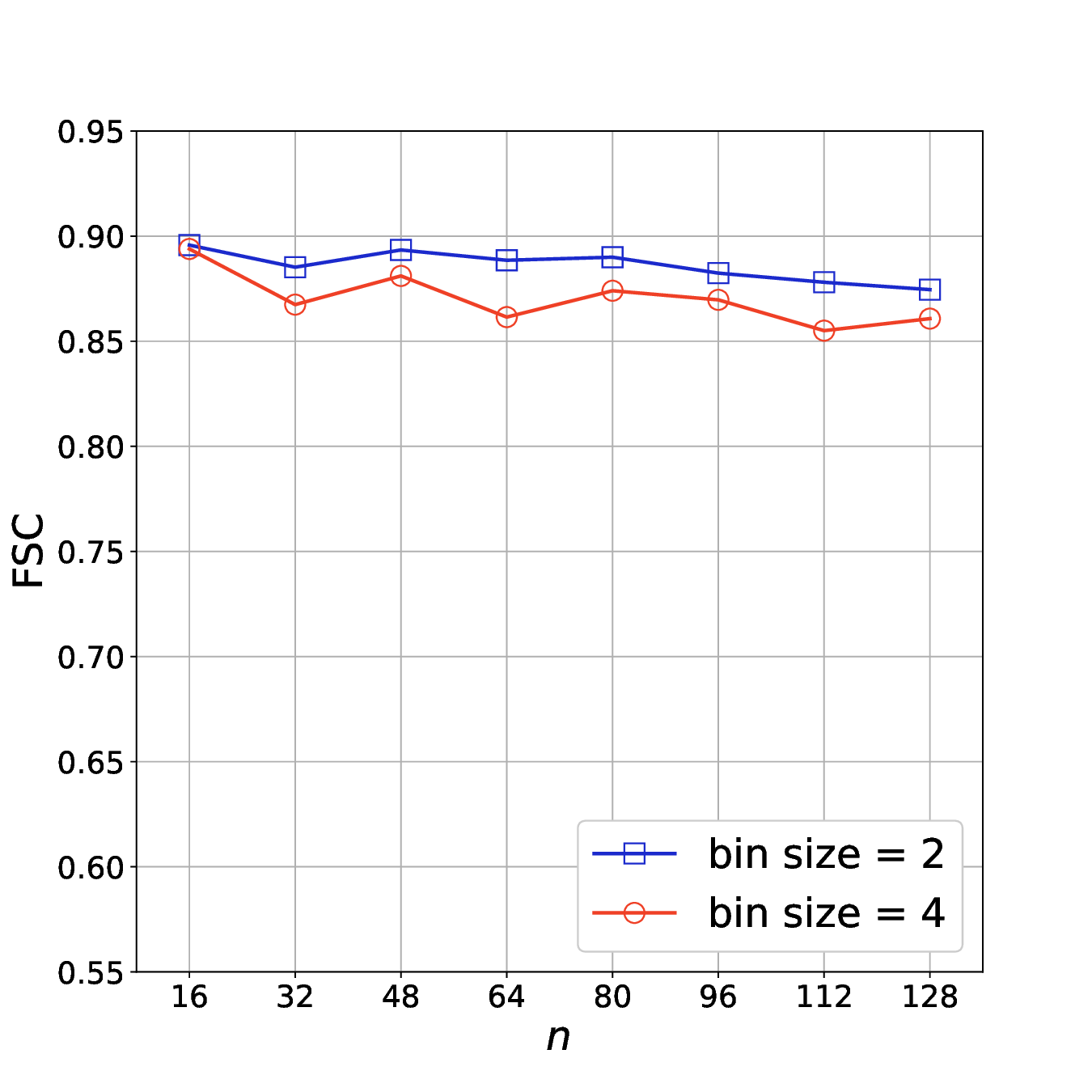}}
\subfigure[$m$]{
\label{fig:para:subfig:m} 
\includegraphics[width=3in]{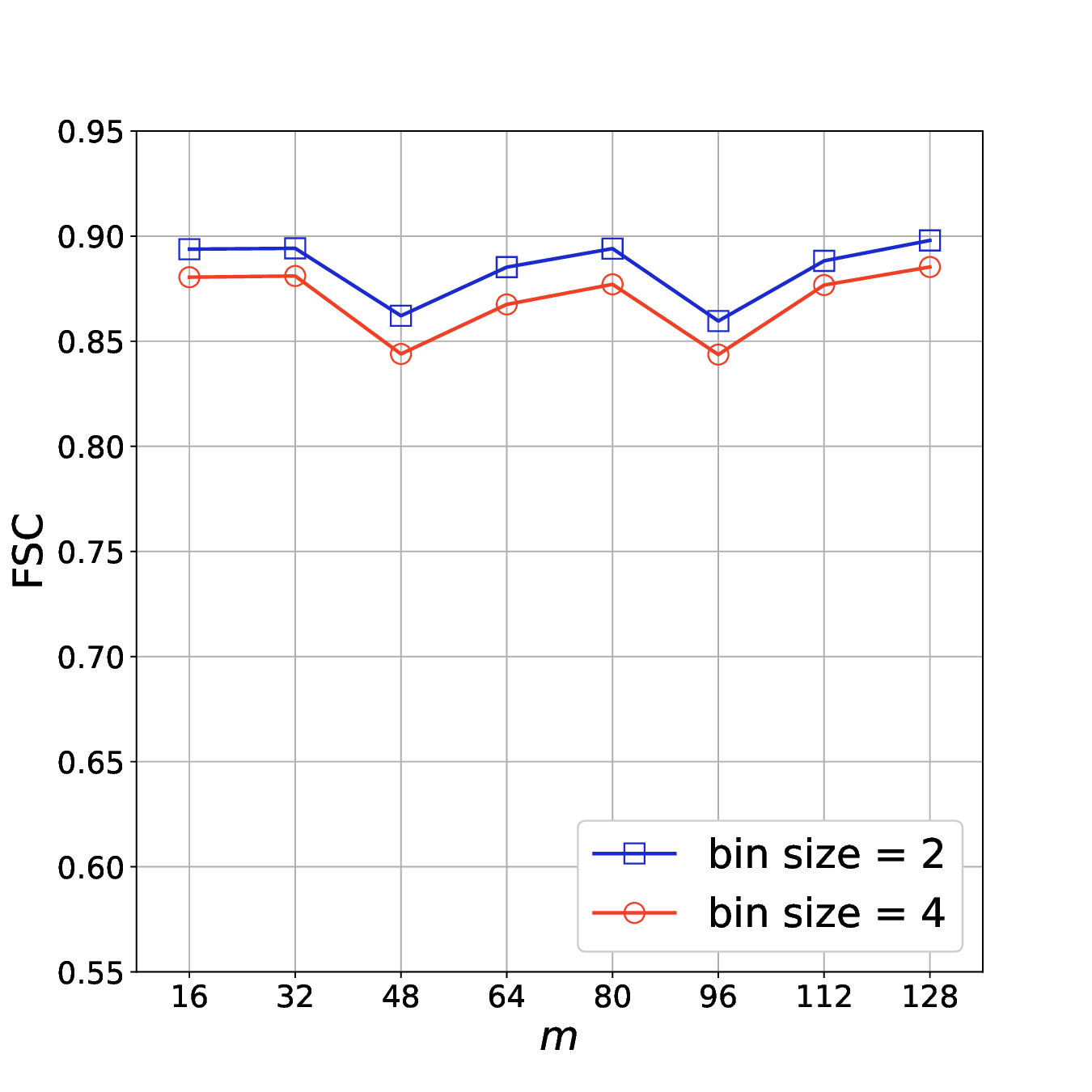}}
\subfigure[batch size]{
\label{fig:para:subfig:batch} 
\includegraphics[width=3in]{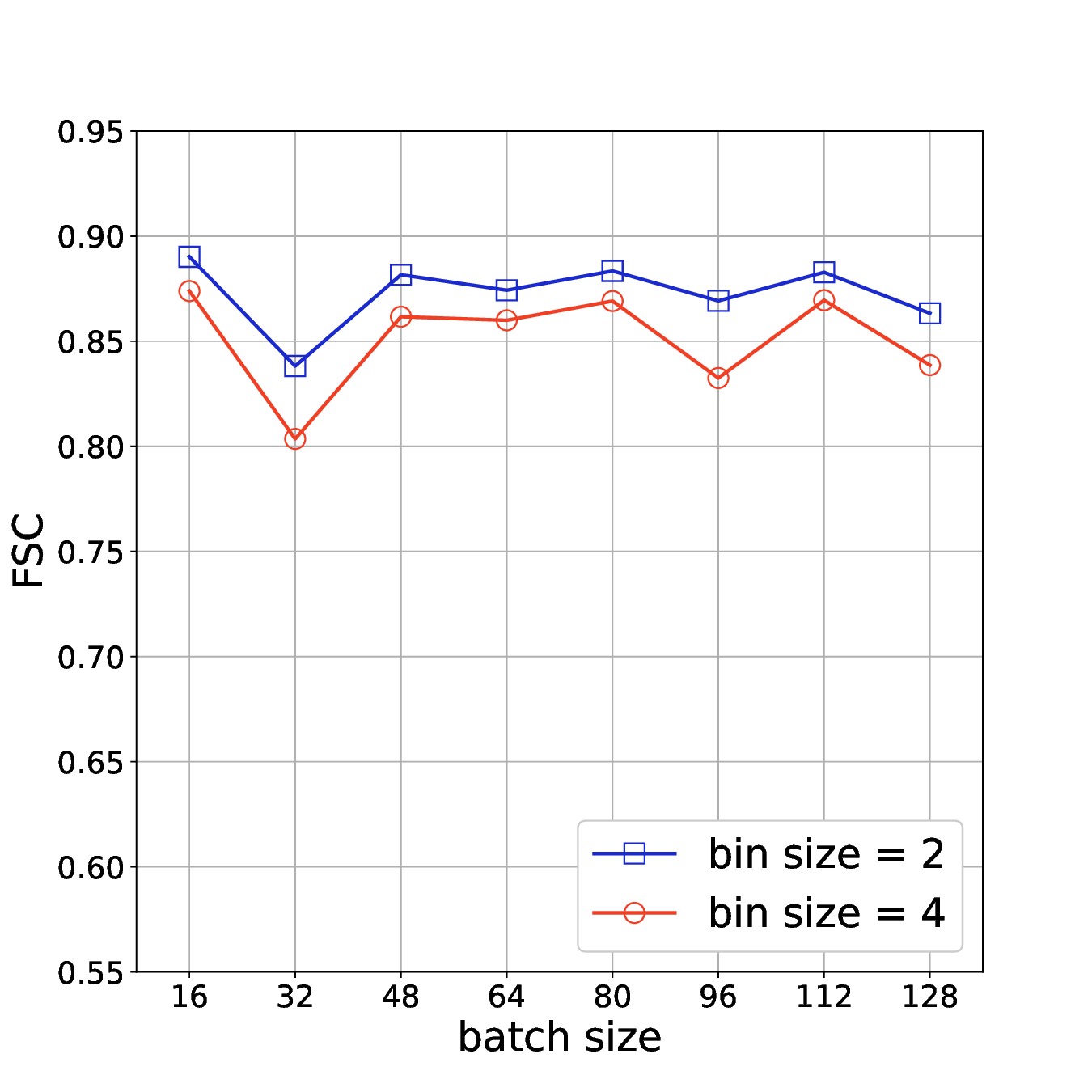}}
\subfigure[epoch number]{
\label{fig:para:subfig:epoch} 
\includegraphics[width=3in]{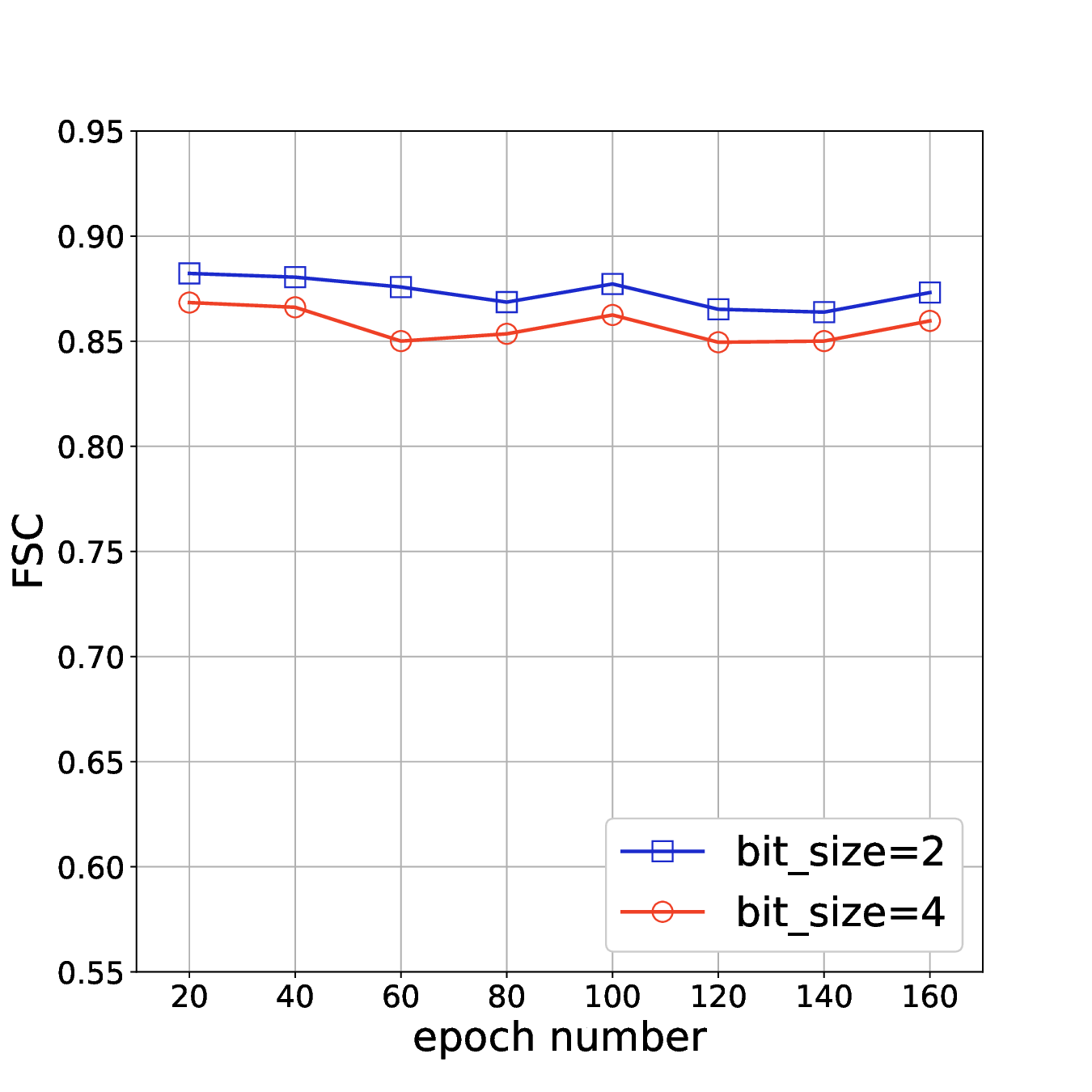}}
\caption{The sensitivity of SNEFY-LDL with regard to the four hyperparameters: $n$, $m$, batch size and epoch number.}
\label{fig:para} %% label for entire figure
\end{figure*}
\end{document}